%% file: main.tex
\documentclass[12pt, onecolumn, draftclsnofoot, peerreview]{IEEEtran}%

\usepackage{cite}

\ifCLASSINFOpdf
   \usepackage[pdftex]{graphicx}
   \graphicspath{{../pdf/}{../jpeg/}}
   \DeclareGraphicsExtensions{.pdf,.jpeg,.eps}
\else
   \usepackage[dvips]{graphicx}
   \graphicspath{{../eps/}}
   \DeclareGraphicsExtensions{.eps}
\fi
\usepackage{amsmath}
\usepackage{amssymb}
\usepackage{algorithm}
\usepackage{algpseudocode, tabularx}
\usepackage{booktabs}
\usepackage{multirow}
\usepackage{cellspace}

\usepackage{stackengine}

\newlength{\continueindent}
\makeatletter
\newcommand{\multiline}[1]{%
  \begin{tabularx}{\dimexpr\linewidth-\ALG@thistlm}[t]{@{}X@{}}
    #1
  \end{tabularx}
}
\makeatother

\usepackage{array}

\input{definitions}

\begin{document}
%
\title{Progressive Modality Cooperation for Multi-Modality Domain Adaptation}
%
%

%

\author{Weichen Zhang, Dong Xu,~\IEEEmembership{Fellow,~IEEE}, Jing Zhang, and Wanli Ouyang,~\IEEEmembership{Senior Member,~IEEE}
\thanks{Weichen Zhang, Dong Xu (corresponding author) and Wanli Ouyang are with the School of Electrical and Information Engineering, The University of Sydney, NSW, Australia.
Jing Zhang is with the College of Software, Beihang University, Beijing, China.
(e-mail: weichen.zhang@sydney.edu.au; dong.xu@sydney.edu.au; wanli.ouyang@sydney.edu.au, zhang\_jing@buaa.edu.cn).}}

\author{Weichen Zhang, Dong Xu,~\IEEEmembership{Fellow,~IEEE}, Jing Zhang, and Wanli Ouyang,~\IEEEmembership{Senior Member,~IEEE}
}
\maketitle

\begin{abstract}
In this work, we propose a new generic multi-modality domain adaptation framework called Progressive Modality Cooperation (PMC) to transfer the knowledge learned from the source domain to the target domain by exploiting multiple modality clues (\eg, RGB and depth) under the multi-modality domain adaptation (MMDA) and the more general multi-modality domain adaptation using privileged information (MMDA-PI) settings. Under the MMDA setting, the samples in both domains have all the modalities. In two newly proposed modules of our PMC, the multiple modalities are cooperated for selecting the reliable pseudo-labeled target samples, which captures the modality-specific information and modality-integrated information, respectively. Under the MMDA-PI setting, some modalities are missing in the target domain. Hence, to better exploit the multi-modality data in the source domain, we further propose the PMC with privileged information (PMC-PI) method by proposing a new multi-modality data generation (MMG) network. MMG generates the missing modalities in the target domain based on the source domain data by considering both domain distribution mismatch and semantics preservation, which are respectively achieved by using adversarial learning  and conditioning on weighted pseudo semantics. Extensive experiments on three image datasets and eight video datasets for various multi-modality cross-domain visual recognition tasks under both MMDA and MMDA-PI settings clearly demonstrate the effectiveness of our proposed PMC framework.

\end{abstract}


\newpage
\section{Introduction}
\IEEEPARstart{W}{ith} the advancement of different sensors (\eg, Microsoft Kinect and iPhone TrueDepth Camera), data with multiple views/modalities can be employed in various computer vision applications \cite{zhang2018binary, peng2019comic, huang2019multi}. Many works~\cite{gupta2014learning,hoffman2016cross,tu2018semantic} including the multi-modality deep learning methods have shown that multiple complementary modalities (\eg, RGB, depth, and optical flow) can be employed to boost the visual recognition performance under the assumption that a large number of labeled multi-modality data are available. However, the labeled multi-modality data are still scarce in practice. An alternative way is to leverage the off-the-shelf labeled multi-modality data from a related source domain to learn the model for the domain of interest (\ie, the target domain). When compared to the source domain data, the target domain data may be from different illumination conditions and viewpoints, and may be captured by different sensors, leading to the so-called data distribution mismatch issue between the source and target domains. Unfortunately, most of the existing multi-modality deep learning methods \cite{gupta2014learning,hoffman2016cross,tu2018semantic} cannot explicitly deal with this issue. Meanwhile, the existing domain adaptation methods \cite{duan2012domain, duan2012domainkernel,tzeng2014deep, ganin2016domain,zhang2019domain} are mostly designed to mitigate the domain gap by assuming both the source and the target domains only contain the single modality data. Considering the data distribution mismatch from multiple modalities, some simple and straightforward approaches (\eg, the early/late fusion strategies) may not be effective, and in fact may be potentially detrimental.

\begin{figure}[!t]
\vspace{-1.5mm}
\begin{center}
\includegraphics[width=0.9\linewidth]{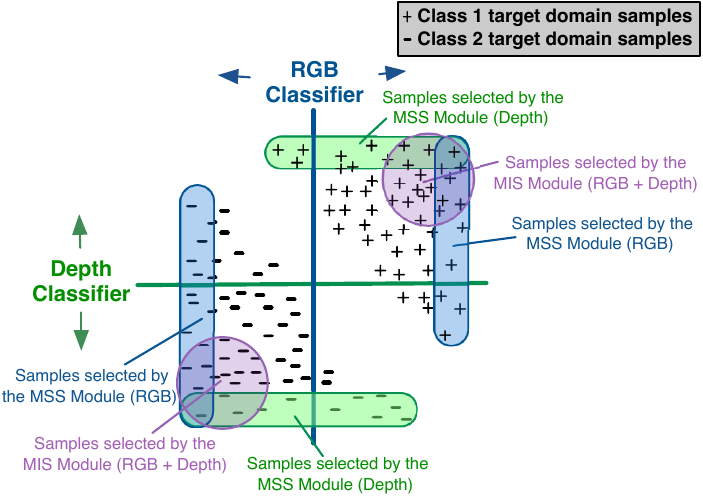}
\end{center}
\vspace{-5mm}
\caption{The motivation of our PMC framework. Two modalities (\ie, the RGB and depth modalities) are used for better illustration. ``+'' and ``-'' indicate the samples from two different classes. 
The modality-specific sample selection (MSS) module selects confident samples based on the prediction scores from each individual modality, while the modality-integrated sample selection (MIS) module selects confident samples based on the prediction scores from all modalities.
During the training process, the number of selected samples from both modules increases progressively. By using both MSS and MIS modules, our PMC jointly captures both modality-complementary and modality-shared information for the target domain.} 
\label{fig:motivation}
\vspace{-4mm}
\end{figure}

Additionally, in many real-world applications, some modalities could be missing in the target domain, which further enlarge the domain distribution mismatch. For example, the training source domain data can be the images (with both RGB and depth modalities) captured by the Kinect camera, while the testing data in the target domain may be web photos or webcam images (with the RGB modality only). 
The previous works have shown that it is helpful to utilize privileged information \cite{vapnik2009new, sharmanska2013learning, ding2014latent, hoffman2016learning, niu2016exploiting}, 
where the additional features from different modalities in the training data can still be useful for classifying the testing data. 
Unfortunately, most of them cannot explicitly deal with the domain distribution mismatch issue, as these works assume that the training data and the testing data are from the same distribution. Considering domain discrepancy, it still remains an open problem on how to utilize the additional modality in the source domain or generate the complete multi-modality target data when some modalities are missing in the target domain. Empirically, by generating the missing modality data in the target domain, some potentially useful information from the generated modalities can still be extracted to improve the classification performance on the target domain.

To this end, we propose a new framework called Progressive Modality Cooperation (PMC) to tackle not only the multi-modality domain adaptation (\textbf{MMDA}) task, where the samples from both source and target domains contain multiple modalities, but also the multi-modality domain adaptation with privileged information (\textbf{MMDA-PI}) task, where some modalities are missing in the target domain.

Under the \textbf{MMDA} setting, the proposed PMC framework aims at better cooperation among multiple modalities by exploiting both modality-specific and modality-integrated information, which is motivated by the consensus and complementary principles in the multi-view learning approaches \cite{xu2013survey}. Specifically, we first train the domain-invariant and modality-specific models based on the labeled source domain data and unlabeled target domain data. Then, to learn more discriminative target domain features, the models are refined based on the target samples with pseudo labels. Motivated by self-paced learning \cite{kumar2010self}, more confident/easier target samples are selected in the early iterations while less confident/harder target samples are selected at the subsequent iterations. 

Additionally, different from the single-modality domain adaptation methods using pseudo labels~\cite{saito2017asymmetric,zhang2018collaborative}, we propose a new sample selection strategy based on the observation that different modalities of the data can  effectively cooperate with each other to achieve better domain adaptation performance. 
In our PMC method, two sample selection modules, namely modality-specific sample selection (MSS) and modality-integrated sample selection (MIS) are then proposed to progressively select and reweigh the target domain samples for improving each modality-specific model.
As shown in Figure \ref{fig:motivation}, the MSS module captures more complementary information from different modalities, namely different samples may be better distinguished by using appearance information from RGB data or using the 3D structure information from depth data. On the other hand, we follow the consensus principle and use the MIS module to learn more modality-shared information (e.g., object shapes in both modalities) by selecting the samples that the models from all the modalities agree on their prediction results, which ensures the reliability of the selected target samples. By simultaneously employing the MSS module and the MIS module in the proposed PMC framework, various domain adaptation methods (\eg, DANN \cite{ganin2016domain}) can be readily incorporated and multiple (\ie,$\ge$ 2) modalities can be effectively integrated. 
  
Under the \textbf{MMDA-PI} setting, the PMC method cannot be directly employed, because some modalities are missing in the target domain. Motivated by learning using privileged information (LUPI) \cite{hoffman2016learning, niu2016exploiting}
, a possible solution is to generate the missing modality for data in the target domain, such that the potentially useful information from the generated modalities can then be extracted to further improve the performance of the subsequent recognition/detection tasks.
Though several domain adaptive modality generation methods \cite{nath2018adadepth,zheng2018t2net} have been proposed, these methods are not specifically designed for both missing modality generation and the subsequent visual recognition tasks. Hence, we further propose a new domain adaptive Multi-Modality data Generation (MMG) method to generate the missing modalities for our PMC. In MMG, the domain distribution mismatch issue is considered by taking advantage of adversarial learning and the semantic information is also preserved by conditioning on the (pseudo) class labels to boost the subsequent classification/detection tasks.  

Overall, our main contributions can be summarized as follows:

(1) Our newly proposed progressive modality cooperation (PMC) framework progressively exploits complementary and consensus information from multiple modalities under both MMDA and MMDA-PI settings. To the best of our knowledge, this is the first deep domain adaptation framework that can handle both settings for various visual recognition tasks. Our PMC method can also effectively cope with the samples with more than two modalities (\eg, 4 modalities), which is not studied in the existing domain adaptation works.

(2) We propose a new modality cooperation strategy by introducing two sample selection modules based on self-paced learning, which respectively select reliable samples based on modality-specific and modality-integrated information in an easy-to-hard fashion.

(3) We propose a multi-modality data generation module (MMG) to generate the missing modality by not only considering domain distribution mismatch but also preserving semantical information in the target domain. The generation process of MMG incorporates the progressively improved pseudo semantic information from our PMC, which can also be readily used in other missing modality generation approaches.

(4) Extensive experiments on various multi-modality cross-domain visual recognition tasks (\eg, image-based object recognition, video-based action recognition and detection) verify the effectiveness, robustness and generalization ability of our framework under both MMDA and MMDA-PI settings.

\vspace{-2mm}
\section{Related Work}
\vspace{-1mm}
\subsection{Domain Adaptation}
\vspace{-1mm}
In recent years, domain adaptation \cite{duan2012domainkernel, duan2012domain, baktashmotlagh2013unsupervised,fernando2013unsupervised,gong2012geodesic, huang2007correcting, bruzzone2010domain, li2014learning, li2018domain,long2015learning, tzeng2014deep,sun2016deep,long2017jan,long2018conditional,pinheiro2018unsupervised, ganin2015unsupervised,ganin2016domain,bousmalis2016domain, hoffman2017cycada, zhang2018collaborative, saito2018maximum, zhang2019domain, zhang2020self} has been widely investigated in computer vision. The conventional domain adaptation methods can be roughly categorized as feature based approaches \cite{baktashmotlagh2013unsupervised,fernando2013unsupervised,gong2012geodesic}, and classifier based approaches \cite{duan2011visual,duan2012domainkernel,duan2012domain,xu2014exploiting,li2014learning, bruzzone2010domain,li2018domain}. 
The deep transfer learning approaches can be roughly categorized as statistic moments-based approaches~\cite{tzeng2014deep,sun2016deep,long2015learning, long2017jan,pinheiro2018unsupervised}, which learn transferable features by employing statistic moments-based regularizers, and adversarial learning-based approaches~\cite{ganin2015unsupervised,ganin2016domain,bousmalis2016domain, hoffman2017cycada, zhang2018collaborative, saito2018maximum, zhang2019domain, zhang2020self}, which use Generative Adversarial Networks (GANs)~\cite{goodfellow2014generative} to learn domain-invariant representation through adversarial learning. In addition, several domain adaptation methods \cite{saito2017asymmetric,zhang2018collaborative, pan2019transferrable, zhang2020self} have been proposed to select pseudo-labeled target samples or prototypes by using different thresholds for different models. Furthermore, Curriculum Learning~\cite{bengio2009curriculum} and Self-Paced Learning~\cite{kumar2010self} are shown to be effective for domain adaptation~\cite{zhang2017curriculum,zou2018unsupervised,zhang2018collaborative,zhang2020self} by progressively selecting pseudo-labeled training samples in an easy-to-hard fashion. More details are provided in the recent survey paper \cite{zhang2019recent}.

However, all these methods mainly focus on how to adapt the models for single modality data, while our PMC method transfers the model based on multi-modality data by exploiting complementary and consensus information from all the modalities.

\vspace{-3mm}
\subsection{Multi-Modality Domain Adaptation}
Recently, a few multi-modality visual recognition applications \cite{gupta2014learning,hoffman2016cross,su2019improving,su2020progressive} have been studied. To address the domain distribution mismatch problem, several multi-modality approaches \cite{niu2015multi,ma2019deep,qi2018unified,munro2019multi} have been proposed for domain adaptation. 
In contrast to these methods \cite{niu2015multi,ma2019deep,qi2018unified,munro2019multi}, our method uses a modality-cooperation scheme with aid of pseudo-labeled target samples to learn more discriminative target domain features based on both modality-specific and modality-shared information by exploiting both complementarity and consistency of multiple modalities under both MMDA and MMDA-PI settings.

\vspace{-3mm}
\subsection{Learning Using Privileged Information}
When one or more views of data in the target domain are missing (\ie, the MMDA-PI setting), our method is also related to learning using privileged information (LUPI) \cite{vapnik2009new,sharmanska2013learning,ding2014latent,hoffman2016learning}, where the additional features available at the training stage become unavailable at the testing stage.
However, most LUPI methods assume the training and testing data have the same data distribution.
To deal with the missing modality, some recent low-rank transfer learning approaches \cite{ding2015missing,li2018visual} have been proposed to reduce the data distribution mismatch between the training and testing data by learning a projection matrix from multi-modality data to single-modality data. Nevertheless, these non-deep-learning approaches cannot fully exploit complementary information among different modalities. 

In contrast, we propose a new deep learning-based method to simultaneously deal with domain distribution mismatch and encourage modality cooperation among multi-modality data from both domains by generating missing modalities for the data in the target domain.



\vspace{-3mm}
\subsection{Domain Adaptation for Depth Estimation}
The supervised learning methods (\textit{e.g.,}\cite{saxena2006learning,liu2015deep}) have been intensively used for the monocular depth estimation task over the past decade.
To deal with the domain distribution mismatch, researchers have recently paid more attention to the domain adaptive depth estimation task. Most of these domain adaptive depth estimation approaches \cite{atapour2018real,zheng2018t2net,zhao2019geometry} focus on how to translate the style of input images to the target domain for depth estimation, which are not specifically designed for the subsequent visual recognition tasks. 

Motivated by Conditional GAN\cite{mirza2014conditional}, our newly proposed multi-modality data generation approach additionally uses the gradually improved pseudo category information to progressively generate semantics preserved depth maps for the subsequent visual recognition tasks, which improve the performance of both missing modality generation and the subsequent visual recognition tasks for the target domain. 
The idea of introducing semantic knowledge into the generation process in our approach can also be readily incorporated into these advanced approaches \cite{atapour2018real,zheng2018t2net,zhao2019geometry} for generating high-quality depth images for the subsequent visual recognition tasks.


\begin{figure*}[t]
\begin{center}
\includegraphics[width=0.9\linewidth]{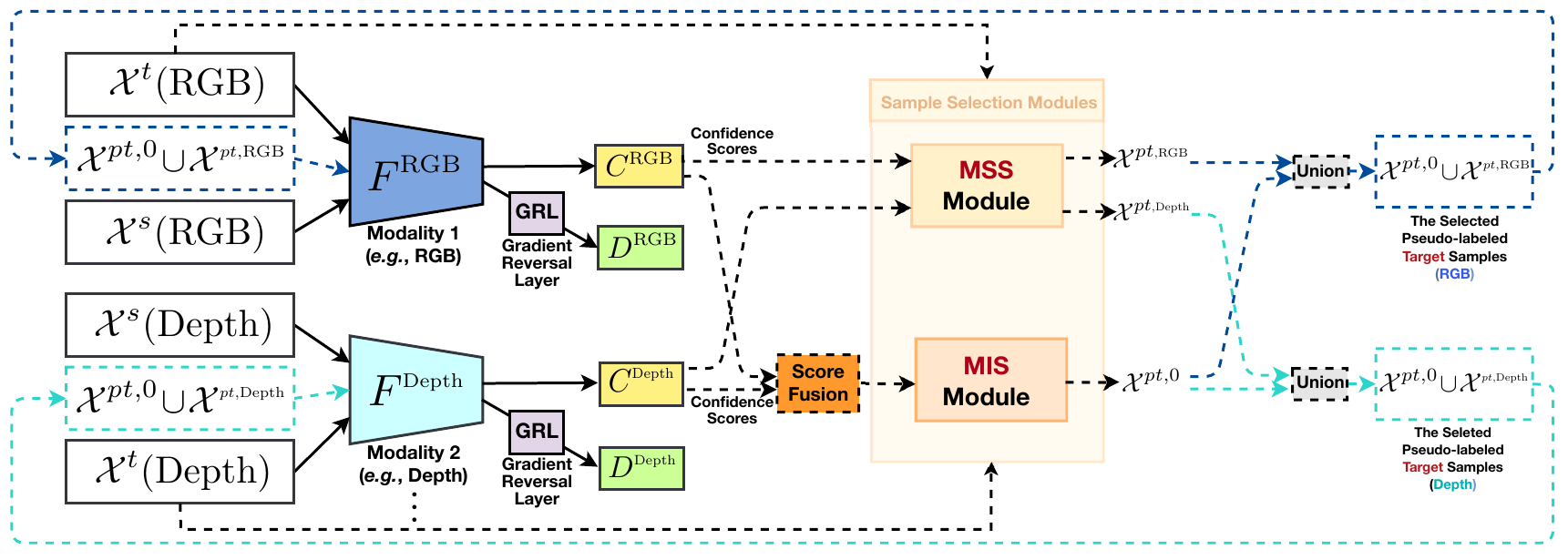}
\vspace{-7mm}
\end{center}
\caption{The architecture of our Progressive Modality Cooperation (PMC) framework. We take the two-view recognition task with both RGB and depth modalities as an example for better illustration. $F^{\text{RGB}}$/$F^{\text{Depth}}$, $C^{\text{RGB}}$/$C^{\text{Depth}}$ and $D^{\text{RGB}}$/$D^{\text{Depth}}$ denote the feature extractor, the image/video classifier and the domain classifier for the RGB/Depth modality, respectively. $\mathcal{X}^{s}(\text{RGB})$/$\mathcal{X}^{s}(\text{Depth})$ and $\mathcal{X}^{t}(\text{RGB})$/$\mathcal{X}^{t}(\text{Depth})$ denote the RGB/Depth modality of the source and target samples, respectively. $\mathcal{X}^{pt,\text{RGB}}$/$\mathcal{X}^{pt,\text{Depth}}$ denotes the pseudo-labeled target samples selected by the Modality-Specific sample Selection (MSS) module based on the RGB/Depth modality. $\mathcal{X}^{pt,0}$ denotes the pseudo-labeled target samples selected by the Modality-Integrated sample Selection (MSS) module. The RGB/depth branch is initially trained by using the domain adversarial learning strategy as in~\cite{ganin2016domain}. Then, for better cooperation among multiple modalities, we improve the model for each RGB/depth branch by additionally using the pseudo-labeled target samples after combining the samples selected by the MSS module and the MIS module, in order to jointly exploit both complementary and consensus information. The dashed lines represent the newly proposed sample selection and re-training procedures.}
\label{fig:framework}
\vspace{-4mm}
\end{figure*}

\vspace{-1mm}
\section{Methodology}
\label{sec:MMDA}
In this section, we introduce our proposed method in detail. 
Formally, let us denote $\mathcal{X}^{s}\!=\!\{(\mathbf{x}^{m}_{i}|_{m=1}^{M^s}, y_i)|_{i=1}^{N^{s}}\}$ as the labeled multi-modality data in the source domain, where $N^{s}$ is the total number of source samples, $M^s$ is the number of modalities in the source domain and $\x^{m}_{i}$ is the $m$-th modality data of the $i$-th labeled source sample.  The corresponding label of the $i$-th source sample is denoted as $y_i$.
Similarly, we denote $\mathcal{X}^{t}\!=\!\{(\x^{m}_{i}|_{m=1}^{M^t})|_{i=N^{s}+1}^{N^{s}+N^{t}}\}$
as the unlabeled target data, where $N^{t}$ is the number of target samples, $M^t$ is the number of modalities in the target domain and $\x^{m}_{i}$ is the $m$-th modality of the $i$-th unlabeled target sample. Under the MMDA setting, the number of modalities in the source and target domains are the same (\ie, $M^s\!=\!M^t$), while under the MMDA-PI setting, the number of modalities in the target domain is smaller than that in the source domain (\ie, $M^t\!<\!M^s$). Under the MMDA setting, we use $\mathcal{X} = \{(\x_i^m|_{m=1}^{M}, d_i)|_{i=1}^{N}\}$ 
to denote the training data from both domains, where $N$ is the total number of samples from both domains (\ie, $N = N^s + N^t$), $x_i^m|_{m=1}^{M}$ is the $i$-th sample with all $M$ modalities (\ie, $M = M^s = M^t$). $d_i \in \{0, 1\}$ is the domain label for the $i$-th sample, where we have $d_i=0$ for the samples in the source domain, and $d_i = 1$ for the samples in the target domain. Under the MMDA-PI setting, we also use $\mathcal{X}$ to denote the training data from both domain, in which the missing modality of the target data can be generated by our MMG network (See Section \ref{sec:mmg}).

\vspace{-3mm}
\subsection{Domain-invariant Feature Learning}
To learn domain-invariant features for each modality, our work builds upon a simple but effective domain adaptation method, domain adversarial neural network (DANN) \cite{ganin2016domain}. Note that any other single-modality domain-invariant feature learning methods can be readily used. Hence, for the $m$-th modality, we train a feature extraction network $F^m( \cdot ; \bt_F^m)$, an image/video classifier $C^m( \cdot ; \bt_C^m)$ and a domain classifier $D^m( \cdot ; \bt_D^m)$, where $\bt_F^m$, $\bt_C^m$ and $\bt_D^m$ are the parameters for the corresponding networks, respectively. The data from different modalities are fed into the corresponding network in our PMC. 


To learn domain invariant features for $F^m$, the adversarial learning strategy \cite{ganin2016domain} is used by confusing the domain classifier $D^m$ with the Gradient Reversal Layer (GRL). The loss $\mathcal{L}_{adv}^m$ used for domain invariant feature learning for the $m$-th modality is defined as follows,
\vspace{-1mm}
\begin{equation}
\label{eqn:dann}
\!\!\!\mathcal{L}_{adv}^m=\frac{1}{N}\sum_{i=1}^N\cL_{D}\left(D^m(F^m(\x_i^m; \bt_F^m); \bt_D^m), d_i\right),\!\!\!
\end{equation}
where $\cL_{D}$ is the binary cross-entropy loss. As shown in \cite{ganin2016domain}, the domain classifier $D^m(\cdot;\bt_D^m)$ is optimized by minimizing the loss $\mathcal{L}_{adv}^{m}$ such that the samples from different domains can be better distinguished. 
While the network $F^m(\cdot;\bt_F^m)$ is optimized by maximizing the objective $\mathcal{L}_{adv}^{m}$ to learn the domain-invariant features.

In addition, in order to learn more discriminative features for the image/video classification task, we also minimize the loss $\mathcal{L}_{src}^{m}$ for learning the image/video classifier $C^m(\cdot;\bt_C^m)$ for the $m$-th modality, which is defined as follows,
\begin{eqnarray}
\vspace{-3mm}
\label{eqn:src}
\mathcal{L}_{src}^{m} = \frac{1}{N^s}\sum_{i=1}^{N^s}\cL_{C}(C^m(F^m(\x^m_i; \bt_F^m);\bt_C^m), y_i),
\vspace{-2mm}
\end{eqnarray}
where $\cL_{C}$ is the cross entropy loss.

The overall objective for our multi-modality domain adaptation framework is to jointly optimize the objective functions from all modalities as follows,
\vspace{-1mm}
\begin{equation}
\label{eqn:danntotal}
\sum_{m=1}^{M} (\min_{\bt_F^m,\bt_C^m}\!\mathcal{L}_{src}^{m}\! +\! \lambda \max_{\bt_F^m}\min_{\bt_D^m} \mathcal{L}_{adv}^{m}),
\vspace{-1mm}
\end{equation}
where $\lambda$ is a trade-off parameter as in \cite{ganin2016domain}.

\vspace{-4mm}
\subsection{Progressive Modality Cooperation}
\vspace{-1mm}
After learning the domain-invariant models for different modalities, we can simply employ the late fusion strategy or concatenate the mid-level features to produce the final prediction results. However, in these simple approaches, the information from different modalities cannot be effectively integrated and the useful information in the target domain is not fully exploited.
As a result, in our proposed method, the highly confident target samples are selected as pseudo-labeled data to improve the recognition performance and learn more discriminative features for the target domain samples. Specifically,
we propose a modality-specific sample selection (\textbf{MSS}) module and a modality-integrated sample selection (\textbf{MIS}) module to facilitate the cooperation between different modalities. The selected samples from the two modules are then combined for improving the models collaboratively. The architecture of our PMC framework is shown in Figure \ref{fig:framework}.

\subsubsection{Modality-Specific sample Selection (MSS) module}
~\\
\indent 
The proposed MSS module aims to enhance the models by using the pseudo-labeled target samples selected from each modality-specific model, such that the complementary information captured by different modalities is exploited. To reduce domain distribution mismatch, we propose to progressively select highly-reliable target samples in an easy-to-hard fashion similar to curriculum learning \cite{bengio2009curriculum} and self-paced learning \cite{kumar2010self}. Then, in order to reduce the detrimental effect by using the wrongly-predicted target samples with incorrect pseudo labels, we additionally re-weight the selected target samples according to their prediction confidence scores. This selection strategy is similar as that in our previous work \cite{zhang2020self} for single modality domain adaptation. 

Specifically, for each modality, let us denote $\{p_c(\x^{m}_{i})|_{c=1}^{N_{c}}\}$ as the output from the softmax layer of the corresponding image/video classifier $C^m(\cdot ; \bt_C^m)$, which represents the predicted class probabilities based on the $m$-th modality of the samples, and $N_{c}$ is the total number of categories. 
The pseudo-label $\tilde{y}^{m}_{i}$ of the data $\x^{m}_{i}$ can be obtained by choosing the category with the highest probability, \textit{i.e.}, $\tilde{y}^{m}_{i} = \arg\max_{c}p_{c}(\x^{m}_{i})$, with its \textit{classification confidence score} as $p_{\tilde{y}^{m}_{i}}(\x^{m}_{i})$.

Then, for the $m$-th modality, we use the classification confidence scores $p_{\tilde{y}^{m}_{i}}(\x^{m}_i)$ as the guidance to sort the target samples at each training epoch.
We select the first $r^m\%$ of pseudo-labeled target samples with the high classification confidence scores to re-train the corresponding branch related to the $m$-th modality, where
$r^m\!\in\![0,100]\!$ is the \textit{individual proportion parameter} representing the percentage of the selected pseudo-labeled target samples, which is automatically decided from the training progress. To ensure quality of the selected pseudo-labeled target samples, for each individual modality at each epoch, $r^m$ is adjusted based on the image/video classification accuracy in the source domain. If the average accuracy from the $m$-th model drop in two consecutive epochs, $r^m$ is automatically reduced by a certain percentage from that in the previous epoch instead of continuously being increased, which follows our previous work SPCAN.

Formally, for the proportion parameter $r^m$ in the MSS module, let $A_i^m$ be the average image/video classification accuracy over all the source samples from modality $m$ at the $i$-th epoch,
and $\bar{A}_i^m$ be the average accuracy of $A_i^m$ from modality $m$ in the first $i$ epochs (\ie, $\bar{A}_i^m = \frac{1}{i}\sum_{j=1}^iA_j^m$) of the modality cooperation process.  
For the total number of $E$ training epochs, the proportion parameter $r^m$ at the $e$-th epoch is then adjusted as follows\footnote{Note, for any modality, we have $\bar{A}_1^m=A_1^m$, namely $\eta_1^m=\eta_2^m=1$, $\forall m$.},
\vspace{-1mm}
\begin{eqnarray}
\begin{aligned}
\label{eqn:rem}
r^m = & \;\frac{\sum_{i=1}^{e}\eta_i^m}{E}, \\
\label{eqn:rankm}
\eta_i^m = & \begin{cases}
-1, & \text{if $A_i^m < \bar{A}_i^m$ and $A_{i-1}^m<\bar{A}_{i-1}^m$},\\
1, & \text{otherwise}.
\end{cases}
\vspace{-3mm}
\end{aligned}
\end{eqnarray}
Moreover, considering that the predicted pseudo-labels for the low-confident samples are more likely to be incorrect, for the $m$-th modality of each target sample, we assign the corresponding weight $w^m(\x^{m}_{i})$ according to its classification confidence score, \ie, $w^m(\x^{m}_{i})=p_{\tilde{y}^{m}_{i}}(\x^{m}_{i})$. 
For ease of representation, we denote $\mathcal{X}^{pt, m}$ as the set of selected target samples with their pseudo labels decided by using the MSS module, and also define the indicator function $s^m(\x^{m}_{i})$, which produces $1$ if $\x^{m}_{i}$ is selected and $0$ otherwise. 


\subsubsection{Modality-integrated sample Selection (MIS) module}
~\\
\indent 
The proposed MIS module exploits the consensus information by selecting the highly-reliable target samples that are agreed by multiple modalities for re-training the network.

Specifically, we firstly produce the fused classification confidence score of each target sample by averaging the classification confidence scores from all modalities. In the same way as in the MSS module, we produce the pseudo-label $\tilde{y}^0_{i}$ of the target sample based on all modalities by choosing the category with the highest probability based on the fused classification confidence scores. 
Then, we use the same selection strategy as used in the MSS module to sort the target samples based on the fused classification confidence scores and use the \textit{group proportion parameter} $r^0$ to decide the percentage of the selected pseudo-labeled target samples for retraining the networks related to all modalities.
The image/video classifier weight for the MIS module $w^0(\x_{i}^{m}|^M_{m=1}) = \frac{1}{M}\sum_{m=1}^M w^{m}(\x_{i}^{m})$ is produced by averaging the corresponding weights from all modalities, which is also assigned to the $i$-th target sample selected by the MIS module. 
Similarly, we denote $\mathcal{X}^{pt,0}$ as the set of pseudo-labeled target samples selected by using the MIS module, and $s^0(\x_{i}^{m}|^M_{m=1})$ as the corresponding indicator function for the $i$-th target sample. 

\subsubsection{Modality cooperation}
~\\
\indent 
As shown in Figure \ref{fig:framework}, for each modality of the whole network, we re-train the corresponding branch by using the pseudo-labeled target samples in the set of both $\mathcal{X}^{pt,m}$ and $\mathcal{X}^{pt,0}$ together with their weights.
The loss $\mathcal{L}_{tar}^{m}$ for re-training the image/video classifier $C^m$ and the feature extractor $F^m$ with the selected pseudo-labeled target samples can be written as,
\vspace*{-1mm}
\begin{eqnarray}\small
\label{eqn:target_cI}
\begin{aligned}
\!\!\!\!\!\!\mathcal{L}_{tar}^{m}\!=\!\frac{1}{N^t}\!\sum_{i=N^s+1}^{N^s+N^t}(&s^m(\x_i^m)w^m(\x_i^m)\cL_{C}(C^m(F^m(\x^m_i; \bt_F^m);\bt_C^m), \tilde{y}^{m}_i)\!\!\!\\
&\!\!+s^0(\x_i^{m}|^M_{m=1})w^0(\x_i^{m}|^M_{m=1})\\
&\!\!\times \cL_{C}(C^m(F^m(\x^m_i; \bt_F^m);\bt_C^m), \tilde{y}^0_i)\!\!\!\!\!\!\!\!\!\!\! %
\end{aligned}
\end{eqnarray}

\subsubsection{The Overall PMC Model} 
~\\
\indent 
The overall objective for training the PMC network is to jointly optimize the three losses as follows,
\vspace*{-1mm}
\begin{equation}
\begin{split}
\label{eqn:MMCtotal}
\!\!\!\sum_{m=1}^{M} (\!\min_{\bt_F^m,\bt_C^m}\!(\mathcal{L}_{src}^{m}+ \mathcal{L}_{tar}^{m})\!+\!\lambda \max_{\bt_F^m}\min_{\bt_D^m} \mathcal{L}_{adv}^{m})
\end{split}
\end{equation}
where $\lambda$ is the trade-off parameter as in Eq. (\ref{eqn:danntotal}).

\setlength{\textfloatsep}{12pt}
\begin{algorithm}[t]
\begin{algorithmic}[1]
\caption{Progressive Modality Cooperation (PMC)}\label{algo:MMC}
\State{\textbf{Input}: Labeled source samples $\{(\x_{i}^{m}|_{m=1}^{M^s},y_{i})|_{i=1}^{N^s}\}$ with $M^s$ modalities, unlabeled target samples $\{(\x_i^{m}|_{m=1}^{M^t})|_{i=N^s +1}^{N^s+N^t}\}$ with $M^t$ modalities, $M^s=M^t$.}
    \State{For each modality, train an initial domain adaptation model by optimizing Eq. (\ref{eqn:danntotal}).}
\Loop~until~$max\_epoch$~is reached: 
    \State \multiline{For each modality, select the set of pseudo-labeled target samples $\mathcal{X}^{pt,m}$ by using the MSS module. \strut}
    \State \multiline{Select the set of pseudo-labeled target samples $\mathcal{X}^{pt,0}$ by using the MIS module.}
    \State \multiline{For each modality, based on the labeled source samples, unlabeled target samples and the selected pseudo-labeled target samples in the set $\mathcal{X}^{pt,m} \cup \mathcal{X}^{pt,0}$, train the PMC network by optimizing Eq. (\ref{eqn:MMCtotal}). \strut}
\EndLoop
\State{\textbf{Output}: The adapted multi-modality models and the predicted category probabilities of the target samples.
}
\end{algorithmic}
\end{algorithm}

In this way, the models for all the modalities can be improved progressively and collaboratively by learning the complementary and shared features from both MIS and MSS modules, respectively. To produce the final prediction results, the late fusion strategy is used to combine the prediction scores from all modality-specific models. The training procedure of our PMC framework is listed in Algorithm \ref{algo:MMC}.

\vspace{-3mm}
\subsection{PMC with Privileged Information (PMC-PI)}
\vspace{-1mm}
In many real-world applications, source domain data contains multiple modalities (\eg, RGB and depth), while some modalities (\eg, depth) of target samples may be missing. However, the additional modalities (\eg, depth) of source samples can be treated as privileged information for boosting the performance in the target domain, which forms our setting of \textit{multi-modality domain adaptation with privileged information} (\textbf{MMDA-PI}).

\begin{figure}[!t]
\vspace{-4mm}
\begin{center}
\includegraphics[width=\linewidth]{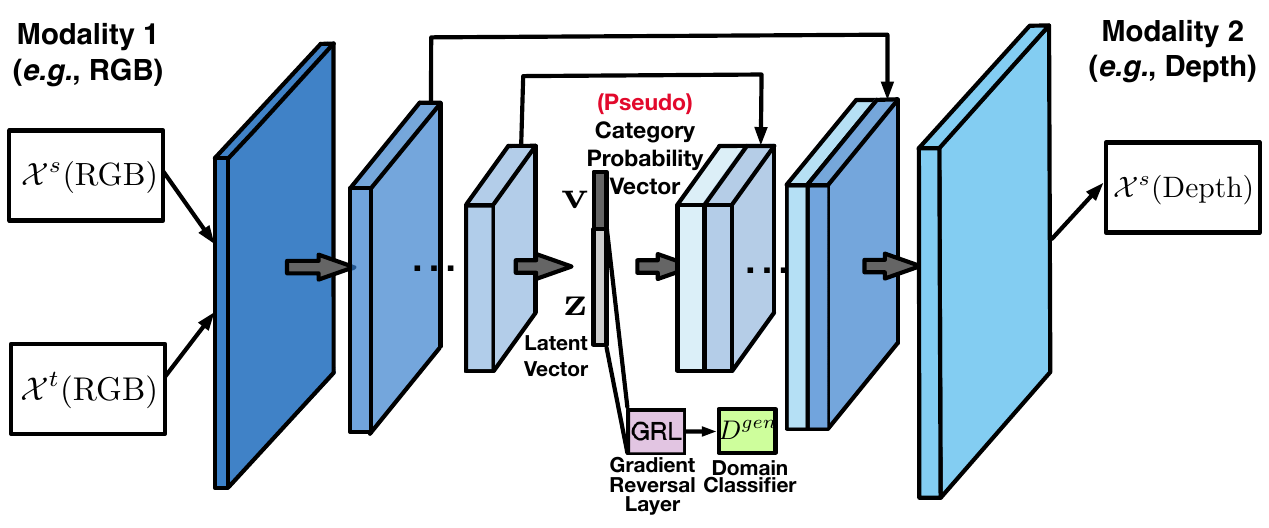}
\end{center}
\vspace{-6mm}
\caption{The architecture of our Multi-Modality Generation (MMG) module. We take the RGB/depth modality as the input/output as an example for better illustration. $\z$ represents the output latent vector from the encoder. $\v$ represents the category probability vector, which is then combined with the latent vector to generate the output depth modality. 
}
\vspace{-2mm}
\label{fig:framework2}
\end{figure}

To address this problem, we propose the Progressive Modality Cooperation with Privileged Information (PMC-PI) method. Specifically, the domain adaptive multi-modality data generation (MMG) module is proposed to generate the missing modality in the target domain for our PMC to further improve the domain adaptation performance. The architecture of our MMG network is illustrated in Figure \ref{fig:framework2}. 

\subsubsection{Multi-Modality Data Generation (MMG)}
\label{sec:mmg}
~\\
\indent 
In contrast to the general domain adaptive data generation approaches \cite{nath2018adadepth,zheng2018t2net,atapour2018real},
the proposed MMG approach takes both domain distribution mismatch and semantic information preservation into consideration.

In detail, in the proposed MMG network, U-Net \cite{ronneberger2015u} is utilized as the basic data generation network. We take the two-modality scenario with RGB and depth modalities as an example, where the depth modality is missing in the target domain. The RGB data (denoted by $\x_{i}^{\text{RGB}}$) and the corresponding ground-truth output depth data (denoted by $\x_{i}^{\text{Depth}}$) from the source domain are used in pairs during the training process. 
Let us denote $E(\cdot; \bt_E)$, $D\!E(\cdot; \bt_{D\!E})$ as the encoder and the decoder networks, where $\bt_E$ and $\bt_{D\!E}$ are the corresponding parameters of the networks, respectively.

If we directly use the generator trained based on the source domain data, we cannot generate the high-quality missing depth data in the target domain due to domain distribution mismatch.
Hence, we propose to learn the domain-invariant features by using the encoder with the domain adversarial learning strategy~\cite{ganin2016domain} and apply a domain classifier $D^{gen}(\cdot;\bt_{D^{gen}})$ after the last layer of the encoder $E(\cdot;\bt_E)$ with the Gradient Reversal Layer (GRL).

The adversarial loss $\mathcal{L}_{adv^{gen}}$ can be formulated as follows,
\vspace{-1mm}
\begin{equation}
\label{eqn:odg}
\mathcal{L}_{adv^{gen}}=\frac{1}{N}\sum_{i=1}^N\cL_{D}\left(D^{gen}(E(\x^{\text{RGB}}_i; \bt_E); \bt_{D^{gen}}), d_i\right)
\end{equation}
where $\cL_D$ is the binary cross-entropy loss and $d_i$ is the domain label for $i$-th sample.

In addition, the category information is not fully exploited in the modality generation process. Inspired by \cite{mirza2014conditional}, our missing modality generator is additionally trained by conditioning on the ground-truth category probability vector of the source domain data. 
The reconstruction loss $\mathcal{L}^{gen}$ of the generator then becomes,
\vspace{-1mm}
\begin{equation}
\label{eqn:genp}
\!\!\mathcal{L}_{gen} = \frac{1}{{N^{s}}}\sum_{i=1}^{N^{s}}\cL_{G}(DE(E(\mathbf{x}_{i}^{\text{RGB}}; \bt_E)\oplus \v_i; \bt_{DE}), \x_{i}^{\text{Depth}})\!\!\!\!\!\!
\end{equation}
\noindent where $\cL_{G}$ is the general $L_1$ loss, $\mathbf{v}_i$ is the one-hot vector based on the category information of the $i$-th source sample and $\oplus$ is the concatenation operation for row vectors to combine the one-hot vector with the latent output vector from the encoder $E(\cdot;\bt_E)$. 


The objective for optimizing the MMG network can be written as follows,
\begin{equation}
\begin{split}
\label{eqn:MMG}
\min_{\bt_{E},\bt_{DE}} \mathcal{L}&_{gen} + \lambda_{gen}\max_{\bt_{E}}\min_{\bt_{D^{gen}}}\mathcal{L}_{adv^{gen}}.\!
\end{split}
\end{equation}

\subsubsection{The Overall PMC-PI Model}
~\\
\indent 
The training procedure of the whole PMC-PI network is listed in Algorithm 2 under the two-modality scenario with RGB and depth modalities. It is worth mentioning that during the training process, our PMC iteratively generates better pseudo labels and pseudo category probability vectors for our MMG to generate better quality missing depth data in the target domain (in step 7), and the newly generated depth data can further strengthen the modality cooperation in our PMC (in step 8).






\setlength{\textfloatsep}{12pt}
\begin{algorithm}[t]
\begin{algorithmic}[1]
\caption{PMC with Privileged Information (PMC-PI)}
\label{algo:PMC-PI}
\State{\textbf{Input}: Labeled source samples $\{(\x_{i}^{\text{RGB}}, \x_{i}^{\text{Depth}},y_{i})|_{i=1}^{N^s}\}$ with both RGB and depth modalities, unlabeled target samples $\{(\x_i^{\text{RGB}})|_{i=N^s +1}^{N^s+N^t}\}$ with the RGB modality only.}
    \State{Train an MMG network by optimizing Eq. (\ref{eqn:MMG}). Freeze the parameters of MMG in the subsequent steps.}
    \State{Train an initial domain adaptation model by optimizing Eq. (\ref{eqn:danntotal}) based on the RGB modality only.} 
    \State{Evaluate the initial domain adaptation model to produce the initial pseudo label and pseudo category probability vector containing the category probabilities of each target sample. }
    \State{Select the pseudo-labeled target samples and construct the set $\mathcal{X}^{pt,\text{RGB}}$ (Note, $\mathcal{X}^{pt,0}=\varnothing, \mathcal{X}^{pt,\text{depth}}=\varnothing$) by using the MSS module based on the RGB modality only. }
\Loop~until~$max\_epoch$~is reached: 
    \State \multiline{Based on the pre-trained MMG, generate the missing depth modality data in the target domain by using the RGB data in the target domain and the latest pseudo category probability vectors of target samples.}
    \State \multiline{Train the PMC network by optimizing Eq. (\ref{eqn:MMCtotal}), where the training samples include both RGB data and the latest generated depth data from the source and target domains together with the selected pseudo-labeled target samples from $\mathcal{X}^{pt,m} \cup \mathcal{X}^{pt,0}$. \strut}
    \State \multiline{Evaluate the learned PMC model from step 8 to produce the pseudo label and pseudo category probability vector containing the category probabilities of each target sample based on both RGB and the latest generated depth modalities. }
    \State \multiline{Select the pseudo-labeled target samples by using the MSS module and the MIS module, which respectively construct the sets $\mathcal{X}^{pt,m}$ and $\mathcal{X}^{pt,0}$. \strut}
\EndLoop
\State{\textbf{Output}: The MMG and PMC models and the predicted probabilities of the target samples.
}
\end{algorithmic}
\end{algorithm}

\vspace{-3mm}
\subsection{PMC for the detection task}
Besides the aforementioned recognition task, our PMC framework can also be readily extended for the detection task (\eg, video action detection). In the video action detection task, our PMC framework uses two Domain Adaptive Faster R-CNN \cite{chen2018domain} networks in parallel as the basic detection network for the RGB and optical flow modalities, respectively. We first briefly review the Domain Adaptive Faster R-CNN method below and then introduce our PMC for the detection task in detail. 

\subsubsection{Domain Adaptive Faster R-CNN} 
~\\
\indent
Domain Adaptive Faster R-CNN is an object detection method that considers the data distribution mismatch between different domains. It builds upon the widely used two-stage detector named Faster R-CNN. Faster R-CNN consists of three major components: a shared feature extractor, a region proposal network (RPN), and a region-of-interest (ROI) based detection head. The shared feature extractor consists of multiple 2D/3D convolutional layers to extract the features and determine the proposals by using the RPN as well as predict the detection boxes by using the detection head. The RPN uses multiple anchors to generate a set of candidate object proposals, which are then used to crop the feature map from the feature extractor to produce the region-of-interest (ROI). The ROI-based detection head takes all ROIs as the input and finally uses a classifier and a regressor to classify and refine the cropped areas (\ie, the bounding boxes) based on the L1 regression loss and the general cross-entropy loss, respectively.


Domain Adaptive Faster R-CNN further introduces two components for domain adaptation, which align the feature distributions at both image-level and the instance-level. Additionally, a consistency regularization term is used to constrain the image-level domain losses and the instance-level domain losses. 
The overall loss of Domain Adaptive Faster R-CNN combines the losses of Faster R-CNN and the image-level domain adversarial loss, the instance-level domain adversarial loss, and the consistency regularization loss.
Please refer to Domain Adaptive Faster R-CNN \cite{chen2018domain} for more details.

\subsubsection{Modality cooperation for Action Detection}
~\\
\indent 
In our PMC framework, based on the aforementioned Domain Adaptive Faster R-CNN method, we use the MSS and MIS modules to select the reliable target bounding boxes as pseudo-labeled samples to retrain the networks for multiple modalities, which is similar as in the recognition task. 

For the video-based action detection task, the detection heads of all modalities are used to predict the bounding boxes, their class labels and the corresponding confidence scores. Then, we use the proportion parameters $r^m$ and $r^0$ to respectively select the samples with high-confident bounding boxes and labels for the MSS and MIS modules in the detection task as the pseudo-labeled detection boxes, respectively.

Specifically, in the MSS module, we first use the detection heads in the $m$-th modality to predict the bounding boxes and their class labels for all the target domain samples. Then, we apply non-maximum suppression (NMS) on all the detected bounding boxes to produce better bounding boxes for each modality. Finally, we select $r^m$\% bounding boxes from the remaining detected bounding boxes with the highest classification confidence scores from the classifier of the detection head of the $m$-th modality as the pseudo-labeled detection boxes to re-train the models of the $m$-th modality. In the MIS module, we first aggregate all the detected bounding boxes and their class labels predicted from all the modalities. Then, we apply NMS on all the detected bounding boxes from all modalities and evaluate these remaining bounding boxes to produce the classification confidence scores by using all the detection heads. Finally, we calculate the fused classification confidence scores by averaging the classification confidence scores from all the modalities and select $r^0$\% remaining bounding boxes after the NMS process that have the highest fused classification confidence scores as the pseudo-labeled detection boxes. These bounding boxes selected by the MIS module are then used to re-train the models from all the modalities. The retraining process of PMC for the detection task is the same as that for the recognition task, except that the bounding boxes of samples are selected for modality cooperation instead of the entire samples.

\vspace{-2mm}
\section{Experiments}
\label{sec:EXP}
In this section, we conduct extensive experiments for three visual recognition tasks (\ie, image-based object recognition, video-based action recognition and video-based action detection) under both MMDA and MMDA-PI settings.

\vspace{-3mm}
\subsection{Datasets}
For the object recognition task, we evaluate our PMC framework by using three datasets, including Washington RGB-D Object dataset \cite{lai2011large}, Bremen RGB-D Object dataset \cite{patricia2017deep} and Caltech-256 dataset \cite{griffin2007caltech}. Following the settings in \cite{li2018visual}, we select the ten common classes\footnote{The ten common classes are Ball, Calculator, Cereal Box, Coffee Mug, Keyboard, Flashlight,  Light Bulb,  Mushroom, Soda Can and Tomato.} from all three datasets.
We refer to these three subsets as Washington-10, Bremen-10 and Caltech-10, which contain 927 and 1053 images with both RGB and Depth modalities, and 1132 images with only RGB modality, respectively.

For the action recognition task, we use four RGB video subsets and two RGB-D video subsets. Two of the RGB video subsets UCF50-6 and Olympic-6 are constructed by using the videos from the six common categories\footnote{The six selected categories include: Basketball, Clean and Jerk, Discus Throw, Diving, Pole Vault and Tennis.} in the UCF50 \cite{reddy2013recognizing} and Olympic Sports\cite{niebles2010modeling} datasets. Another two of the RGB video subsets UCF101-10 and HMDB51-10 are constructed by using the videos from the ten common categories\footnote{The ten selected categories include: Archery(Shoot Bow), Basketball(Shoot Ball), Biking(Ride Bike), Diving, Fencing, Golf Swing(Golf), Horse Riding(Ride Horse), Pull Ups, Punch and Push Ups. Different from the 12 selected classes in \cite{chen2019temporal}, we observe the walk/climb class in HMDB and the walkWithDog/RopeClimbing class in UCF are semantically different.} in the UCF101 \cite{soomro2012ucf101} and HMDB51 \cite{kuehne2011hmdb} datasets. 
The two RGB-D video subsets Berkley-8 and UWA-8 are constructed by using the videos from eight common categories\footnote{The eight selected classes include: Bending, Jump, Jumping Jack, One Hand Wave, Sit Down, Stand Up, Two Hand Punch, and Two Hand Wave.} from BerkeleyMHAD \cite{ofli2013berkeley} and UWA3D \cite{rahmani2016histogram} datasets, respectively. 

For the action detection task, we use two RGB video datasets with labeled bounding boxes. Due to the lack of labeled samples in the video datasets, we only construct two subsets (\ie, UCF24-2 and JHMDB21-2) with the videos from two common categories (\ie, basketball and golf) that are respectively selected from two benchmark datasets UCF-101-24 \cite{soomro2012ucf101} and J-HMDB-21 \cite{jhuang2013towards} for action detection. UCF24-2 and JHMDB21-2 contain 273 videos and 86 videos, respectively. 

\vspace{-3mm}
\subsection{Implementation Details}
The proposed method is implemented in the Pytorch framework.
For the image-based object recognition task, the RGB and depth images are resized and cropped to 224$\times$224. The depth maps are further normalized to the values of 0-255, then colorized with the Colorjet method in \cite{eitel2015multimodal}.
We use the ResNet-18 \cite{he2016deep} model pre-trained on ImageNet as the backbone network for all the deep learning-based methods including ours.
The image classifier and the domain classifier for both modalities consist of two fully-connected layers, which are added after the last layer of the feature extractor (\ie, the 18-$th$ layer for ResNet-18). 
The learning rates for the image classifiers and the domain classifiers are set as 10 times of the backbone network, which are initially set as $0.0015$.
As in DANN \cite{ganin2016domain}, the same INV learning rate decreasing strategy and adaptation factor are used. Mini-batch stochastic gradient descent is used for optimization. The batch size, momentum, and weight decay are set as 16, 0.9 and $3\times10^{-4}$, respectively. We set the trade-off parameters $\lambda$ = 0.3, and $\lambda_{gen}$ = 0.1.

For the video-based action recognition task, we re-scale the frames from all modalities into the size of 340$\times$256 and produce the frame-based data in the same way as in \cite{wang2018temporal}. 
To capture more temporal information in videos, we further exploit the optical flow maps between two consecutive RGB/depth frames to form the additional modalities (\ie, Flow/DepthFlow), which are both generated by using the TVL1 optical flow extraction algorithm \cite{zach2007duality}. We use the BN-Inception \cite{ioffe2015batch} model pre-trained on ImageNet as the backbone network for both our method and all re-implemented deep learning-based baseline methods. We use the same training strategy and classifier structures as those in the image classification task except that we set the initial learning rate as 0.001 and the batch size as 48.

For the video action detection task, we use the I3D features \cite{carreira2017quo} pre-trained on Kinetics for all modalities and use the Faster RCNN method \cite{ren2015faster} as the basic framework. We exploit the optical flow maps as in the action recognition task to form the optical flow modality. The mini-batch sizes to train the RPN and the detection head are 256 and 512, respectively. We set the initial learning rate as 0.01, and use the same INV learning rate decrease strategy and adaptation factor as in the image and video recognition tasks. Similar to \cite{chen2018domain}, two domain classifiers are added after the RPN and the detection head for each modality.

\vspace{-3mm}
\subsection{Experiments under the MMDA setting}
\vspace{-1mm}
\subsubsection{Image-based object recognition}
~\\
\indent Under the \textbf{MMDA} setting, the samples from the source and target domains contain both RGB and depth modalities.
Since most existing works are not specifically designed for multi-modality domain adaptation, we compare our work with five supervised multi-modality approaches including (1) HHA \cite{gupta2014learning}, (2) CNN Features \cite{schwarz2015rgb}, (3) Cross-Modal \cite{hoffman2016cross} (4) FusionNet \cite{eitel2015multimodal} and (5) RCFFusion \cite{loghmani2019recurrent}, five single-modality domain adaptation approaches including (6) DAN \cite{long2015learning}, (7) JAN \cite{long2017jan}, and (8) CDAN \cite{long2018conditional}, (9) DANN \cite{ganin2016domain}, and (10) iCAN \cite{zhang2018collaborative}, and two multi-modality domain adaptation approaches including (11) UFMDA \cite{qi2018unified} and (12) MM-SADA \cite{munro2020multi}. For all methods, the input samples from the two domains have both modalities. For all single-modality domain adaptation methods, different networks for all modalities are learned separately and the scores from all networks are fused in a late fusion fashion. Similar to our PMC, the baseline method iCAN \cite{zhang2018collaborative} uses pseudo-labeled target samples for re-training the network. For the multi-modality domain adaptation methods \cite{qi2018unified,munro2020multi} designed for the cross-media retrieval tasks, we modified the classifier of their methods in order to fairly compare them with our PMC method.   

\begin{table}[t]
\setlength\tabcolsep{5pt}\small
\vspace{-2mm}
\caption{Accuracies (\%) of different methods on the Washington-10 (W) and Bremen-10 (B) datasets under the \textbf{MMDA} setting with both RGB and depth modalities for the object recognition task. The single-modality methods use the late fusion strategy by averaging the classification scores from multiple modalities.}
\label{table:MMC_main} 
\begin{center}
\vspace{-2mm}
\resizebox{0.95\linewidth}{!}{%
\begin{tabular}{c c c c}
\toprule
\multirow{2}*{\textbf{}}&\multirow{2}*{\textbf{Methods}}  & B $\rightarrow$ W & W $\rightarrow$ B\\
& & (RGB-D) & (RGB-D) \\
\midrule
\multirow{5}*{\shortstack{\textbf{Multi-modality}\\\textbf{w/o Domain Adaptation}}}&HHA \cite{gupta2014learning}            & 73.5           & 85.7                    \\
& CNN Features \cite{schwarz2015rgb}           &   77.3        &  89.3
     \\
&   Cross-Modal \cite{hoffman2016cross}         &     79.0       &  90.6 \\
& FusionNet \cite{eitel2015multimodal}           &     79.4       &  89.9 \\
& RCFusion \cite{loghmani2019recurrent}           &     81.6       &  92.7 \\
\cmidrule{1-4}
\cmidrule{1-4}
\multirow{5}*{\shortstack{\textbf{Single-modality}\\\textbf{with Domain Adaptation}}}&DAN \cite{long2015learning}           &   87.6       &  98.6   \\
&DANN \cite{ganin2016domain}            & 88.0    & 98.9                     \\
&JAN \cite{long2017jan}           &    90.1    &  99.0 \\
&iCAN \cite{zhang2018collaborative}           &     91.3       &  99.4 \\
&CDAN \cite{long2018conditional}   &      92.1  &   99.5  \\
\cmidrule{1-4}
\multirow{3}*{\shortstack{\textbf{Multi-modality}\\\textbf{with Domain Adaptation}}} & UFMDA \cite{qi2018unified}    &     89.7 & 99.7\\
&  MM-SADA \cite{munro2020multi} &    89.5 & 99.4  \\
&\textbf{PMC(Ours)}            & \textbf{94.1} 
& \textbf{100.0} 
\\
\bottomrule
\end{tabular}
}
\end{center}
\vspace{-3mm}
\end{table}



From Table \ref{table:MMC_main}, we observe that single modality domain adaptation approaches using the late fusion strategy generally outperform the multi-modality approaches without using domain adaptation, which shows it is beneficial to explicitly reduce the data distribution mismatch between the two domains. In addition, one of the baseline multi-modality with domain adaptation approach UFMDA concatenates the mid-level features with the attention module. However, the view-specific statistical property may be ignored by simply concatenating the features. The MM-SADA method \cite{munro2020multi} also extends DANN for the multi-modality tasks by additionally using a modality correspondence classifier for the alignment of different modalities (\ie, RGB and depth).
Nevertheless, the complementary information of each modality in their method is not explicitly considered and learned in the modality-alignment procedure. In contrast, by using our progressive modality cooperation strategy, both consensus and complementary information are well captured in our PMC method. 
We also compare our PMC method with the  TS-SPCAN method proposed in our previous work \cite{zhang2020self}, which achieves 92.0\% and 99.9\% for the B$\rightarrow$W and W$\rightarrow$B tasks, respectively.
Our method PMC thus achieves the best accuracies for both tasks (B$\rightarrow$W and W$\rightarrow$B).

\begin{table}
\setlength\tabcolsep{8pt}\small
\vspace{-2mm}
\begin{center}
\caption{Accuracies (\%) of different methods when using each single modality on the Washington(W) and Bremen(B) datasets under the MMDA setting. The number outside the parentheses is the accuracy after integrating the results from both modalities, and the two numbers inside the parentheses are the results of different methods when using each individual modality (\ie, RGB and depth) before modality fusion.}
\vspace{-2mm}
\resizebox{0.85\linewidth}{!}{%
\label{table:single_da}
\begin{tabular}{c c c}  
\toprule
\multirow{2}*{\textbf{Methods}} & B$\rightarrow$W & W$\rightarrow$B  \\
& RGB-D (RGB, D) & RGB-D (RGB, D) \\
\midrule
ResNet18 \cite{he2016deep}         & \underline{80.2} (79.1, 59.6)  &  \underline{93.3} (92.6, 58.9)  \\
\midrule
DAN \cite{long2015learning}        & \underline{87.6} (86.9, 64.6) & 98.6 (\underline{99.3}, 61.7) \\
DANN \cite{ganin2016domain}         & 88.0 (\underline{88.4}, 65.0) & 98.9 (\underline{99.2}, 64.1) \\
JAN \cite{long2017jan}           &     \underline{90.1} (89.2, 65.6)      &  99.0 (\underline{99.4}, 63.1)
     \\
iCAN \cite{zhang2018collaborative}           &     91.3 (\underline{91.5}, 69.8)       &  99.4 (\underline{99.7}, 67.9)
     \\
CDAN \cite{long2018conditional}   &      \underline{92.1} (91.4, 70.5) &    99.5 (\underline{99.9}, 68.3) \\
MM-SADA \cite{munro2020multi}         &  89.5 (\underline{89.7}, 78.5) &  99.4 (\underline{99.5}, 74.2) \\
PMC(Ours) & \textbf{\underline{94.1}} (\textbf{93.8}, \textbf{92.6}) & \textbf{\underline{100.0}} (\textbf{\underline{100.0}}, \textbf{98.2}) \\
\bottomrule             
\end{tabular}
}
\end{center}
\vspace{-3mm}
\end{table}

To further demonstrate the effectiveness of our modality cooperation strategy, we also compare our method with some baseline methods when evaluating on each single modality. 
From Table \ref{table:single_da}, we observe that our PMC outperforms the baseline methods not only based on the overall results by integrating all modalities but also the individual results from each single modality (especially the depth modality). It is also observed that some baseline methods achieve lower overall results after integrating both modalities when compared with the results from the RGB modality. In contrast, our PMC fully exploits complementary information from the weak modality (\ie, the depth modality), which generally improves the results when using the RGB or the depth modality and achieves better performance after fusing both modalities. 


\subsubsection{Video-based action recognition} 
~\\
\indent 
For the video-based recognition tasks, our method is evaluated on the RGB video datasets with two modalities (\ie, RGB and Flow) and the RGB-D video datasets with four different modalities (\ie, RGB, Depth, Flow, and DepthFlow). Similar as in the object recognition task, we compare our work with several methods, including two supervised multi-modality approaches
(1) TSN \cite{wang2018temporal} and (2) Cross-Modal \cite{hoffman2016cross}; two single-modality domain adaptation approaches (3) DANN \cite{ganin2016domain} and (4) iCAN \cite{zhang2018collaborative}; one multi-modality domain adaptation approach (5) UFMDA \cite{qi2018unified}; one two-stream cross-dataset action recognition approach (6) MM-SADA \cite{munro2020multi}; and two 3D ConvNet-based cross-dataset action recognition approaches (7) DAAA \cite{Jamal2018ddaa}, and (8) TA$\!^3$N \cite{chen2019temporal}.
We follow the training and evaluation protocols in \cite{wang2018temporal} for all the methods.  From the results in Table \ref{table:MMC_four_compare}, our PMC method outperforms all the state-of-the-art methods in all six cases, which again shows the effectiveness of our PMC. 

\renewcommand*{\thefootnote}{\fnsymbol{footnote}}

\begin{table}[t]
\setlength\tabcolsep{3pt}
\vspace{-2mm}
\small
\caption{Accuracies (\%) of different methods on the UCF50-6 (U6), Olympic-6 (O6), UCF101-10 (U10), HMDB51-10 (H10), Berkeley-8 (B8) and UWA-8 (U8) datasets under the \textbf{MMDA} setting. Berkeley-8 and UWA-8 contain both RGB and depth frames. }
\label{table:MMC_four_compare}
\begin{center}
\vspace{-6mm}
\resizebox{\linewidth}{!}{%
\begin{tabular}{c | c c c c | c c }
\toprule
\footnotesize{\multirow{2}*{\textbf{Methods}}}  
& \footnotesize{U6$\rightarrow$O6}& \footnotesize{O6$\rightarrow$U6} & \footnotesize{U10$\rightarrow$H10} & \footnotesize{H10$\rightarrow$U10} & \multicolumn{2}{c}{\footnotesize{B8$\rightarrow$U8} \,\,\,  \footnotesize{U8$\rightarrow$B8}}\\& \multicolumn{4}{c|}{(RGB-Flow)} & \multicolumn{2}{c}{(RGB-D-Flow-DFlow)}\\
\midrule
TSN \cite{wang2018temporal} & 93.9 &    88.9     & 80.5 & 90.1  & \multicolumn{2}{c}{64.8 \;\;\;\;\;\;\  54.3}\\
Cross-Modal \cite{hoffman2016cross} & 90.6 & 89.7 & 79.7 & 90.4 & \multicolumn{2}{c}{ 76.0 \;\;\;\;\;\;\  70.8} \\
DAAA \cite{Jamal2018ddaa}  & 91.6 & 90.0 & - & -  &  \multicolumn{2}{c}{- \;\;\;\;\;\;\,\;\;\;\;\, -}          \\
DANN \cite{ganin2016domain} & 94.3 &  91.0   & 82.6 & 91.7 &\multicolumn{2}{c}{ 75.6 \;\;\;\;\;\;\  74.3} \\ 
iCAN \cite{zhang2018collaborative} & 94.7 & 91.8  & 83.4 & 92.8 & \multicolumn{2}{c}{88.3 \;\;\;\;\;\;\ 83.1 } \\
UFMDA \cite{qi2018unified} & 95.3 & 90.7 & 82.9 & 92.1 & \multicolumn{2}{c}{82.1 \;\;\;\;\;\;\  75.9} \\
MM-SADA \cite{munro2020multi}         &  96.1 &  91.7 & 83.8 & 92.7 & \multicolumn{2}{c}{81.9 \;\;\;\;\;\;\  78.2} \\
TA$\!^3$N \cite{chen2019temporal} & 98.2 & 92.9 & - & - & \multicolumn{2}{c}{- \;\;\;\;\;\;\,\;\;\;\;\,  -} \\
\textbf{PMC(Ours)} & \textbf{98.4} & \textbf{94.8}   & \textbf{89.7}& \textbf{98.1} & \multicolumn{2}{c}{\textbf{98.7} \;\;\;\;\;\;\  \textbf{92.4}} \\
\bottomrule
\end{tabular}
}
\end{center}
\vspace{-2mm}
\end{table}



\subsubsection{Video-based action detection} 
~\\
\indent 
We also evaluate our methods PMC for the video-based action detection task. We compare our work with several deep learning methods, including two supervised detection methods (1)\,Faster R-CNN+FPN\,\cite{lin2017feature} and (2)\,PCSC\,\cite{su2019improving}; and two single-modality domain adaptation approaches (3)\,DA Faster R-CNN-s \cite{chen2018domain} and (4)\,DA Faster R-CNN\,\cite{chen2018domain}. DA Faster R-CNN-s is a special case of DA Faster R-CNN, which uses Faster R-CNN \cite{ren2015faster} as the base detector and only applies a gradient reversal layer after the last feature extraction layer to extract the domain invariant features in the same way as in DANN \cite{ganin2016domain} for the recognition task. For the single-modality domain adaptation approaches, we combine the bounding boxes from the two modalities and then perform NMS. All the methods use the I3D \cite{carreira2017quo} features, 
and are evaluated with the IoU threshold $\delta = 0.5$. 
The results in Table\,\ref{table:PMC_det} indicate that our PMC method performs the best for both cases.

\begin{table}[t]
\vspace{-2mm} 
\setlength\tabcolsep{8pt}
\caption{Results (mAPs \% at the frame level) of different methods on the UCF24-2 (U2) and JHMDB21-2 (J2) datasets under the \textbf{MMDA} setting (IoU threshold $\delta=0.5$). }
\label{table:PMC_det}
\begin{center}
\vspace{-3mm}
\resizebox{0.75\linewidth}{!}{%
\begin{tabular}{c c c c }
\toprule
\multirow{2}*{\textbf{Methods}} & U2$\rightarrow$J2& J2$\rightarrow$U2  \\
& RGB-Flow & RGB-Flow \\
\midrule
Faster R-CNN + FPN \cite{lin2017feature}  & 69.1 &   55.9 \\
PCSC \cite{su2019improving} & 71.3 & 59.7 \\
DA Faster R-CNN-s \cite{chen2018domain}  & 77.6 &  71.2 \\
DA Faster R-CNN \cite{chen2018domain} & 78.9 & 73.8    \\ 
\textbf{PMC(Ours)} & \textbf{83.1} & \textbf{77.3} \\
\bottomrule
\end{tabular}
}
\end{center}
\vspace{-2mm}
\end{table}

\subsubsection{Ablation study of our PMC Framework}
~\\
\indent 
We take the B$\rightarrow$W task with two modalities as an example to further investigate the effectiveness of our PMC method.
Table \ref{table:dammc_abl} reports the results of DANN \cite{ganin2016domain}, which is the baseline method without using pseudo-labeled target samples and two variants of our method:
(1) \textbf{PMC w/o MIS} is the variant of PMC without the MIS module.
(2) \textbf{PMC w/o MSS} is the variant of PMC without the MSS module.

\begin{table}[t]
\vspace{-2mm} 
\setlength\tabcolsep{8pt}\small
\caption{Accuracies (\%) of DANN and two variants of our PMC as well as our method PMC for the B$\rightarrow$W task under the \textbf{MMDA} setting.}
\vspace{-6mm}
\label{table:dammc_abl}
\begin{center}
\resizebox{\linewidth}{!}{%
\begin{tabular}{ccccc}  
\toprule
\textbf{Methods} & DANN \cite{ganin2016domain} & PMC w/o MIS & PMC w/o MSS & PMC \\
\midrule
B$\rightarrow$W   & 88.0  &  91.8  & 93.6 & 94.1 \\
\bottomrule             
\end{tabular}
}
\end{center}
\vspace{-2mm}
\end{table}

From the results in Table \ref{table:dammc_abl}, we have the following observations. First, the method ``PMC w/o MIS'' outperforms DANN by a large margin, which verifies that better representation of each modality can be learned by exploiting the complementary information from different modalities. Secondly, the method ``PMC w/o MSS'' outperforms ``PMC w/o MIS'', which suggests that the proposed integration scheme (\ie, MIS) can effectively select more reliable target samples by discovering the consensus information. Thirdly, if both MSS and MIS modules are used, PMC achieves the best results.

\begin{figure}\scriptsize
\vspace{-2mm}
\begin{center}
\includegraphics[width=\linewidth]{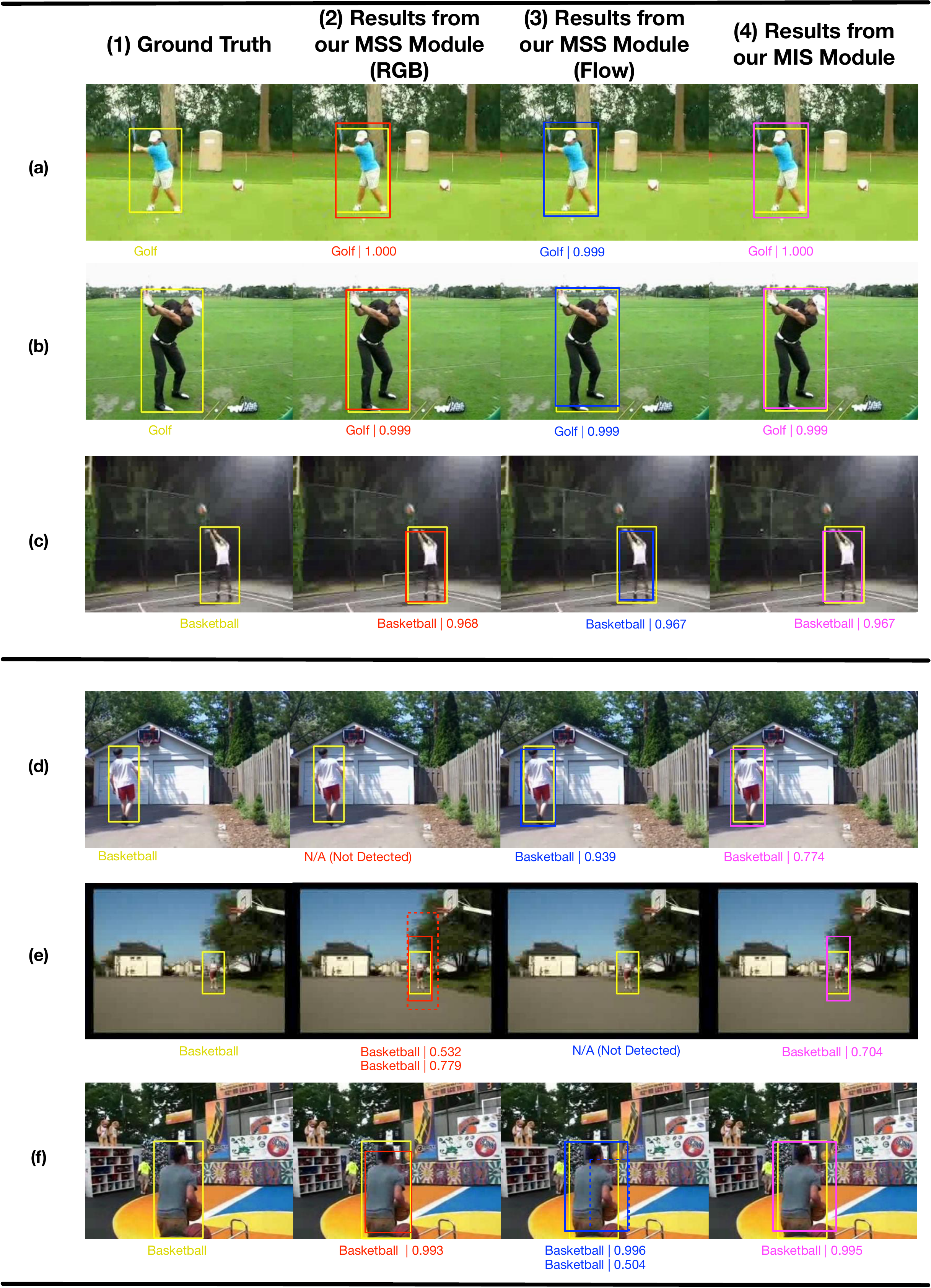}
\end{center}
\vspace{-4mm}
\caption{Visualization of the selected detection bounding boxes by using the MSS and MIS modules. The yellow boxes indicate the ground truth bounding boxes of the video frames. The selected bounding boxes and the corresponding class labels and confidence scores under each image are provided. In (a), (b), (c), the bounding boxes are correctly predicted by using both MSS and MIS modules. In (d), (e), (f), the incorrectly predicted bounding boxes by using the detection head of each modality are shown in the dash lines, which include the missing bounding boxes (\ie, the 2nd column in (d), and the 3rd column in (e)) and the wrongly predicted bounding boxes that do not have sufficient overlap with the ground-truth bounding boxes (See the 2nd column in (e), and the 3rd column in (f)). However, after fusing the bounding boxes from the two modalities and performing Non-Maximum Suppression(NMS), we can still obtain the correctly predicted bounding boxes by using our MIS module (See the 4th column in (d), (e), (f)). }
\label{fig:bbx}
\vspace{-1mm}
\end{figure}

\subsubsection{Visualization of the Selected Detection Boxes in PMC}
~\\
\indent 
For the video-based action detection task, we also visualize some of the selected bounding boxes from our PMC framework by respectively using (1) the MSS module for the RGB modality, (2) the MSS module for the optical flow modality, and (3) the MIS module by integrating both modalities, which correspond to the second, third and fourth columns in Figure \ref{fig:bbx}. We also show the ground truth bounding boxes of the video frames in the first column of Figure \ref{fig:bbx}.

From Figure \ref{fig:bbx}, we have the following observations. Firstly, as shown in Fig. \ref{fig:bbx} (a)-(c), the bounding boxes selected by using the MSS module based on each individual modality (\ie, RGB and optical flow) have large overlapping areas with the ground-truth bounding boxes, and the bounding boxes selected by using the MIS module are also correctly predicted. This demonstrates the effectiveness of our MSS and MIS modules for selecting the correctly predicted bounding boxes.

Secondly, as shown in Fig. \ref{fig:bbx} (d)-(f), in some cases the selected detection boxes are wrongly predicted by using the detection head of each modality. However, after fusing the bounding boxes from the two modalities and performing Non-Maximum Suppression(NMS), we can still produce the correctly predicted bounding boxes by using our MIS module. As a result, the samples with the correctly predicted bounding boxes can still be selected by our MIS module, which demonstrates the effectiveness of our MIS module for integrating the results from both modalities.

By combining the bounding boxes from both MSS and MIS modules, our PMC method can achieve the best performance for the video-based action detection task.

\begin{table}
\setlength\tabcolsep{8pt}
\vspace{-2mm} 
\caption{Accuracies (\%) of our PMC on the W$\rightarrow$B task by setting different hyper-parameter $\alpha$.}
\label{table:sensitivity} 
\begin{center}
\vspace{-2mm}
\resizebox{0.9\linewidth}{!}{%
\begin{tabular}{c c c c c c c}
\toprule
\textbf{$\alpha$} & 0.5 & 0.8 & 1.0 & 1.2 & 1.5 & 2.0\\
\midrule
W$\rightarrow$B & 99.9 & 99.9 & 100.0 & 99.9 & 99.8 & 99.6\\
\bottomrule
\end{tabular}
}
\end{center}
\vspace{-3mm}
\end{table}

\begin{table}
\setlength\tabcolsep{2pt}\small
\begin{center}
\caption{Accuracies (\%)  of our PMC framework when using different combination of four modalities (\ie, RGB (RGB), Depth (D), optical flow (Flow), and depth optical flow (DFlow)) on the Berkeley-8 (B8) and UWA-8 (U8) datasets under the MMDA setting.}
\label{table:MMC_four1}
\vspace{-2mm}
\resizebox{0.98\linewidth}{!}{%
\begin{tabular}{c c c c c c c}
\toprule
\multirow{2}*{\textbf{\# Modalities}} &\multirow{2}*{\textbf{Modality}}  & \multicolumn{2}{c}{B8 $\rightarrow$ U8} & \multicolumn{2}{c}{U8 $\rightarrow$ B8}\\
& & DANN \cite{ganin2016domain} & \textbf{PMC(Ours)} & DANN \cite{ganin2016domain} & \textbf{PMC(Ours)} \\
\midrule
\multirow{4}*{1} & RGB & 50.3 & \textbf{56.9}  & 59.7  & \textbf{71.3}   \\
&D & 63.4 &\textbf{70.1} & 64.6 & \textbf{74.9}  \\
&Flow  & 78.1 & \textbf{87.3}  &  71.8 &\textbf{77.2}  \\
&DFlow & 63.5 & \textbf{75.9} & 66.1 & \textbf{73.6} \\    
\midrule
&RGB-D  & 64.5&  \textbf{78.4}&
72.2&\textbf{78.9}\\
2&RGB-Flow &73.6&\textbf{92.9}&
73.8&\textbf{86.0} \\
&RGB-DFlow  &68.8&\textbf{91.6}&
73.3&\textbf{86.7}\\
\midrule
&RGB-D-Flow &73.1& \textbf{96.9} &
74.1&\textbf{89.9}\\
3&RGB-D-DFlow &70.3& \textbf{93.7} &
73.3&\textbf{89.4}\\
&RGB-Flow-DFlow &72.7& \textbf{95.5}&
74.4&\textbf{90.1}\\
\midrule
4&RGB-D-Flow-DFlow &75.6& \textbf{98.7}&
74.3&\textbf{92.4}\\
\bottomrule
\end{tabular}
}
\end{center}
\vspace{-3mm}
\end{table}

\subsubsection{Analysis of our PMC framework}
\label{sec:param}
~\\
\indent 
During the training progress, the proportion parameters $r^0$ and $r^m$ are automatically adjusted based on the training accuracies of the samples in the source domain because the distributions of the source and target domains still overlap with some extent after adaptation. 
$r^0$ plays a similar role in the MIS module as $r^m$ plays in the MSS module.
Here, we take $r^0$ as an example to investigate the sensitivity of this proportion parameter, in which we set the ratio of the selected pseudo-labeled target samples as $\alpha r^0$ (by multiplying a hyper-parameter $\alpha$). 
In Table \ref{table:sensitivity}, we take the W$\rightarrow$B task under the MMDA setting as an example, in which we set $\alpha$ in the range of
\{0.5, 0.8, 1.0, 1.2, 1.5, 2.0\}. From Table \ref{table:sensitivity}, we observe the accuracy variation is smaller than 0.5\%, which indicates that our method is not sensitive to $r^0$. 



To verify that different modalities can cooperate with each other when applying our PMC method on data with more than two modalities, we further conduct the experiments by using different combination of modalities from \{RGB, Depth, Flow, DepthFlow\} on the Berkeley-8 and the UWA-8 datasets under the MMDA setting. We compare our method with the baseline method DANN \cite{ganin2016domain}. We only show the results when the RGB modality is included as one of the modalities, since it is easier to collect the RGB modality in real-world applications. 
From the results shown in Table \ref{table:MMC_four1}, we have the following observations. 

Firstly, when each single modality is used, our PMC method outperforms the DANN method in all the settings, which verifies that the proposed PMC method extracts more discriminative modality-specific features by gradually selecting more reliable pseudo-labeled target samples.

Secondly, when fusing more modalities, the results of the DANN method are only slightly improved in general and the results are even worse than the counterparts when using each single modality in some cases. For example, for the B8 $\rightarrow$ U8 task, the result of the DANN method using the single optical flow modality is 78.1\%. However, after fusing the optical flow modality with other modalities, such as RGB, the accuracy decreases to 73.6\%. A possible explanation is that the weak modality (\eg, RGB) with worse performance has a detrimental effect on the final results after using the simple late fusion strategy. However, the accuracy of our PMC method using both optical flow and RGB modalities improves from 87.3\% to 92.9 when compared with that of using the optical flow modality only, which indicates different modalities are effectively fused by using our method.

\begin{figure}[!t]
\vspace{-10pt}
\begin{center}
\includegraphics[width=0.95\linewidth]{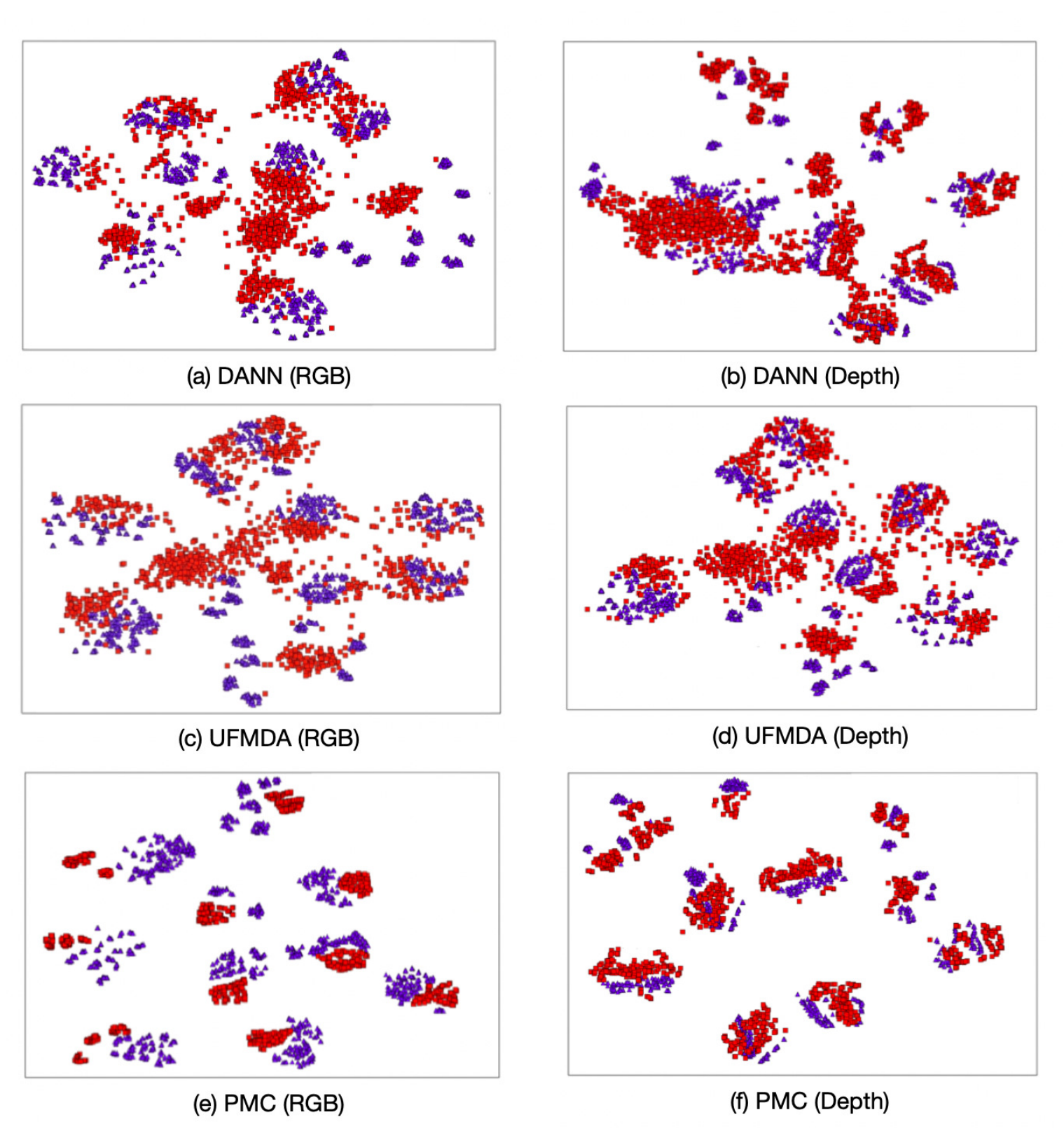}
\end{center}
\vspace{-20pt}
\caption{
T-SNE visualization results for the B$\rightarrow$W task under the MMDA setting. The source and target samples are shown in blue and red, respectively. The results based on the RGB modality (\resp, depth modality) are shown in (a), (c) and (e) (\resp, (b), (d) and (f)). 
}
\vspace{-4pt}
\label{fig:tsne1}
\end{figure}

Furthermore, we visualize the domain discrepancy based on the extracted features from all modalities by using T-SNE. We take the B$\rightarrow$W task as an example to visualize the features of DANN, UFMDA and our PMC using both RGB and depth modalities. From Fig. \ref{fig:tsne1}, our method PMC-PI using either RGB or depth modality can better group the source and target samples from each class, which demonstrates the effectiveness of our progressive modality cooperation strategy and our missing modality generation process.

\vspace{-3mm}
\subsection{PMC under the MMDA-PI setting}
Under the MMDA-PI setting, some modalities of the target data are missing, and in this case, our goal is to exploit the extra modality in the source domain to further boost the recognition performance in the target domain. We evaluate our PMC-PI method for the image-based object recognition task using Washington-10 (W), Bremen-10 (B) and Caltech-10 (C) datasets.
The depth data are assumed to be missing in the target domain, as it is generally harder to collect the depth data than the RGB data in real-world applications.
To the best of our knowledge, the existing deep learning methods are not specifically designed for the MMDA-PI setting.
As a result, we combine our newly proposed MMG sub-network with several deep learning methods as the new baselines, including four single-modality domain adaptation methods (\ie, DAN \cite{long2015learning}, DANN \cite{ganin2016domain}, JAN \cite{long2017jan} and CDAN \cite{long2018conditional}); and two multi-modality domain adaptation methods UFMDA \cite{qi2018unified} and MM-SADA \cite{munro2020multi}, which are referred to as (1) DAN+MMG, (2) DANN+MMG, (3) JAN+MMG, (4) CDAN+MMG, (5) UFMDA+MMG and (6) MM-SADA+MMG in Table \ref{table:PMC_main}, respectively. In addition, for more comprehensive comparison, in Table \ref{table:single_dapi}, we also report the results of all methods by using only single-modality (\eg, RGB or depth) and by using multiple modalities with the late fusion strategy.
For the source samples, we use all modalities (\ie, RGB-D) as the input, while for the target samples we only use the existing modalities (\ie, RGB) as the input. Note, we do not report the result of UFMDA \cite{qi2018unified} in Table \ref{table:single_dapi}, because there is no individual classifier for each modality in UFMDA.

\begin{table}[t]
\setlength\tabcolsep{6pt}\small
\vspace{-2mm}
\caption{Accuracies (\%) of different methods on Washington-10 (W), Bremen-10 (B) and Caltech-10 (C) under the \textbf{MMDA-PI} setting for the object recognition task.}
\label{table:PMC_main}
\begin{center}
\vspace{-3mm}
\resizebox{0.95\linewidth}{!}{%
\begin{tabular}{c c c c c}
\toprule
\textbf{Methods} & B $\rightarrow$ W & W $\rightarrow$ B & W $\rightarrow$ C & B $\rightarrow$ C \\
\midrule
DAN\cite{long2015learning}+MMG           &     86.0       &  97.6
      &     73.9         &  76.5 \\
DANN\cite{ganin2016domain}+MMG      & 87.4           & 98.9           & 76.2           & 80.1           \\
JAN\cite{long2017jan}+MMG      &  89.8         & 98.0 &  78.8 &    81.8      \\
CDAN\cite{long2018conditional}+MMG & 90.6 & 99.3 & 80.3 & 79.7\\
UFMDA\cite{qi2018unified}+MMG  & 88.3  & 98.1 & 79.1 & 80.9  \\
MM-SADA\cite{munro2020multi}+MMG  & 89.3 & 99.0 & 78.9 & 81.2  \\
\textbf{PMC-PI(Ours)}           & \textbf{92.1}           & \textbf{100.0}          & \textbf{81.6}           &  \textbf{82.4}  \\    
\bottomrule
\end{tabular}
}
\end{center}
\vspace{-3mm}
\end{table}

From Table \ref{table:PMC_main}, we observe our PMC-PI method outperforms all the existing baseline methods in all four cases. From the results under the MMDA-PI setting, we have similar observations as those under the MMDA setting. From Table \ref{table:single_dapi}, we also observe that our PMC-PI using each single modality (see the results in parentheses) also achieves the best result when compared to the baseline methods in combination with our MMG, which again demonstrates the effectiveness of our modality cooperation scheme under the MMDA-PI setting. 

\begin{table}
\setlength\tabcolsep{10pt}\small
\vspace{-2mm} 
\begin{center}
\caption{Accuracies (\%) of different methods in combination with our MMG module when using each single modality (RGB or depth) on the Washington(W) and Bremen(B) datasets under the MMDA-PI setting.}
\vspace{-2mm}
\resizebox{0.95\linewidth}{!}{%
\label{table:single_dapi}
\begin{tabular}{c c c}  
\toprule
\multirow{2}*{\textbf{Methods}} & B$\rightarrow$W & W$\rightarrow$B  \\
& RGB-D (RGB, D) & RGB-D (RGB, D) \\
\midrule
DAN \cite{long2015learning} + MMG        & 86.0 (\underline{86.9}, 48.9) & 97.6 (\underline{99.3}, 48.3) \\
DANN \cite{ganin2016domain} + MMG        & 87.4 (\underline{88.4}, 46.8) & 98.9 (\underline{99.2}, 51.4) \\
JAN \cite{long2017jan} + MMG          &     \underline{89.8} (89.2, 51.1)      &  98.0 (\underline{99.4}, 52.7)  \\
CDAN \cite{long2018conditional} + MMG  &      90.6 (\underline{91.4}, 53.6) &    99.3 (\underline{99.8}, 54.1) \\
MM-SADA \cite{munro2020multi} + MMG  &   89.3 (\underline{89.4}, 66.3) & 99.0 (\underline{99.1}, 59.2) \\
\textbf{PMC(Ours)} & \textbf{\underline{92.1}} (\textbf{91.7}, \textbf{90.2}) & \textbf{\underline{100.0}} (\textbf{\underline{100.0}}, \textbf{99.1}) \\
\bottomrule             
\end{tabular}
}
\end{center}
\vspace{-3mm}
\end{table}

\textit{Analysis of our MMG network.} 
We take the most challenging task W$\rightarrow$C as an example to investigate the effectiveness of our MMG network when generating the depth images in the target domain (\ie, Caltech-10) which does not contain the ground-truth depth images.
We conduct three additional experiments by removing the domain classifier from the MMG network (GenD) and/or generating depth images without using the category probability vector (CV).
From Table \ref{table:dammg_abl}, we observe that both ``PMC-PI w/o CV'' and ``PMC-PI w/o GenD'' achieve better results than ``PMC-PI w/o CV and w/o GenD'', which verifies the effectiveness of our CV and GenD modules. By using both modules, our PMC-PI method achieves the best accuracy. 

\begin{table}[t]
\vspace{-2mm} 
\setlength\tabcolsep{10pt}\small
\begin{center}
\caption{Accuracies (\%)  of different variants of our MMG network in PMC on the Washington-10 (W) and Caltech-10 (C) datasets under the MMDA-PI setting. }
\label{table:dammg_abl}
\vspace{-2mm}
\resizebox{0.65\linewidth}{!}{%
\begin{tabular}{c c}
\toprule
\textbf{Methods}                   & W $\rightarrow$ C \\
\midrule
PMC-PI w/o CV and w/o GenD &     79.0 \\
PMC-PI w/o CV & 79.3 \\
PMC-PI w/o GenD & 80.7\\
PMC-PI      & 81.6 \\
\bottomrule    
\end{tabular}
}
\end{center}
\vspace{-3mm}
\end{table}


\begin{figure}
\begin{center}
\includegraphics[width=0.85\linewidth]{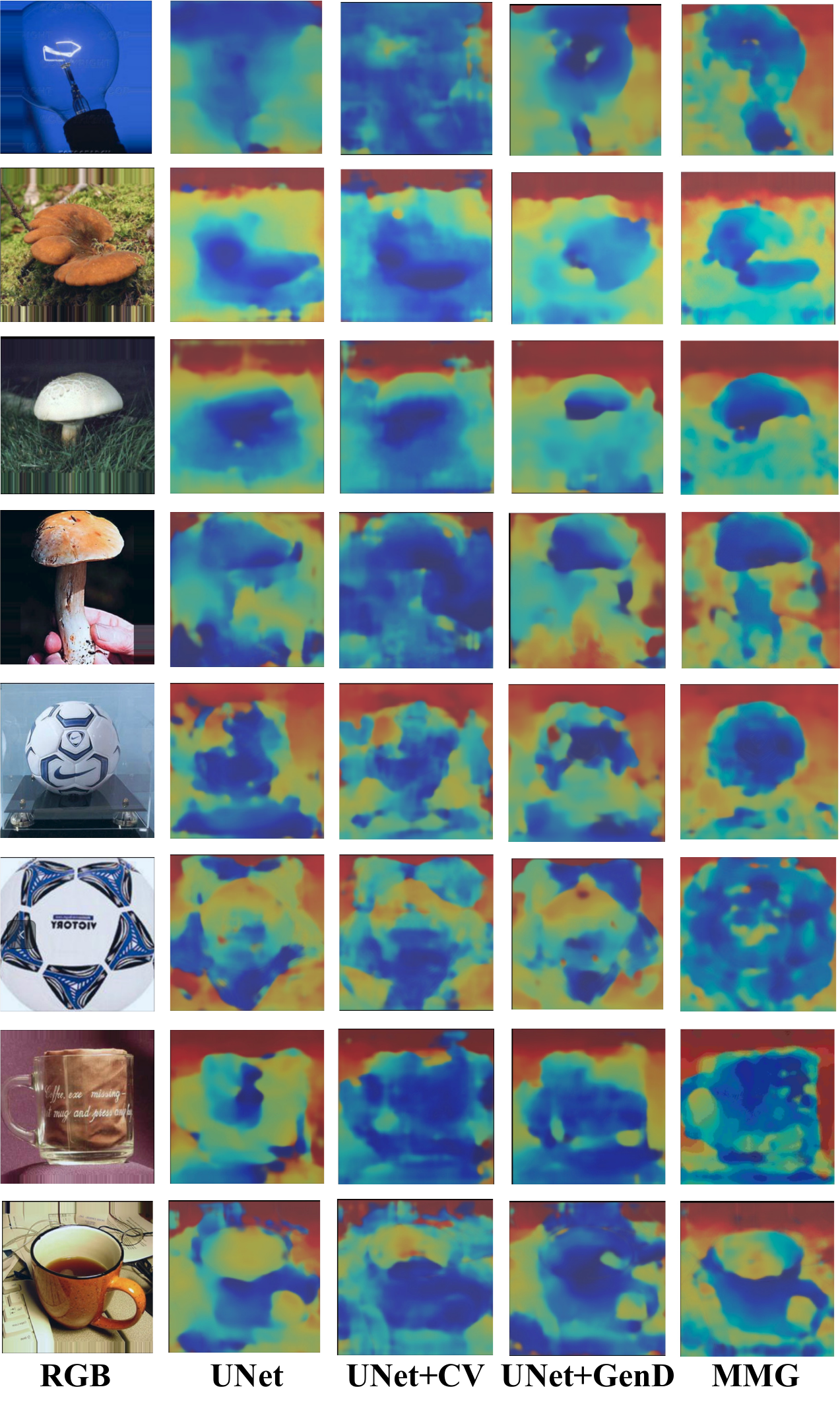}
\vspace{-7mm}
\end{center}
  \caption{Visualization results of the generated depth images. }
\label{fig:vis}
\vspace{-2mm}
\end{figure}

In Figure \ref{fig:vis}, we also visualize the generated depth images in order to intuitively show the effectiveness of our generation network MMG. The images in the five columns correspond to (1) the original RGB images (\ie, RGB), (2) the depth images generated with UNet \cite{ronneberger2015u} only (\ie, UNet), (3) the depth images generated with UNet and by conditioning the depth on the Category probability Vector (\ie, UNet+CV) (4) the depth images generated with UNet and the domain classifier for the generation network (\ie, UNet+GenD), and (5) the depth images generated with our final generation network with both CV and GenD modules (\ie, MMG). It can be observed that our MMG sub-network (\ie, the 5$^{th}$ column) generates the depth images with the best visual quality.


\vspace{-2pt}
\section{Conclusion}
In this work, we have proposed a new framework named Progressive Modality Cooperation (PMC) for the multi-modality domain adaptation task under two different settings MMDA and MMDA-PI. Under the MMDA setting, the samples from both source and target domains contain all the modalities. As a result, the proposed PMC framework takes advantage of both modality-specific and modality-integrated information to select reliable target samples for better cooperation among different modalities. Under the MMDA-PI setting, some modalities of the target samples are missing. To this end, we have proposed a new domain adaptive data generation method MMG to generate the missing modality data for the target domain samples by preserving the semantic information. Comprehensive experiments for image-based object recognition, video-based action recognition and video-based action detection have demonstrated the effectiveness of our proposed PMC framework under both MMDA and MMDA-PI settings.



%



\vspace{-2pt}
\section*{Acknowledgment}
This work was supported by the Australian Research Council (ARC) Future Fellowship under Grant FT180100116 and ARC DP200103223.


\ifCLASSOPTIONcaptionsoff
  \newpage
\fi



%


\vspace{-2pt}
{\small
\bibliographystyle{IEEEtran}
\bibliography{main}
}
%

\newpage
\begin{IEEEbiography}
[{\includegraphics[width=1in,height=1.25in,clip,keepaspectratio]{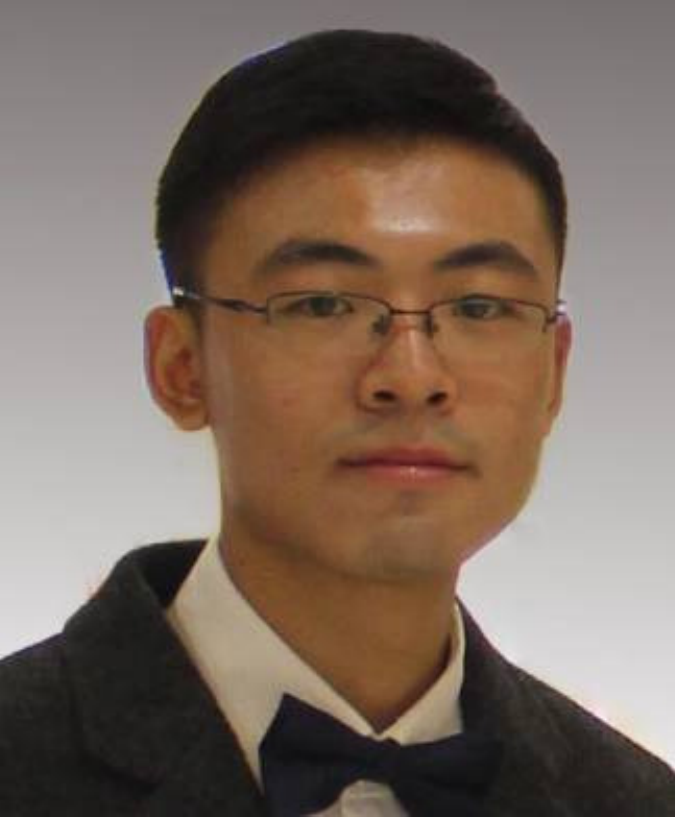}}]
{Weichen Zhang} received the B.S. degree in school of Information Technology from the University of Sydney in 2017. He is currently working toward the PhD degree in the school of Electrical and Information Engineering, the University of Sydney. His current research interests include deep transfer learning and its applications in computer vision.
\end{IEEEbiography}

\begin{IEEEbiography}
[{\includegraphics[width=1in,height=1.25in,clip,keepaspectratio]{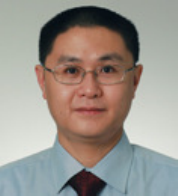}}]
{Dong Xu} received the BE and PhD degrees from University of Science and Technology of China, in 2001 and 2005, respectively. While pursuing the PhD degree, he was an intern with Microsoft Research Asia, Beijing, China, and a research assistant with the Chinese University of Hong Kong, Shatin, Hong Kong, for more than two years. He was a post-doctoral research scientist with Columbia University, New York, NY, for one year. He worked as a faculty member with Nanyang Technological University, Singapore. Currently, he is a professor and chair in Computer Engineering with the School of Electrical and Information Engineering, the University of Sydney, Australia. His current research interests include computer vision, statistical learning, and multimedia content analysis. He was the co-author of a paper that won the Best Student Paper award in the IEEE Conference
on Computer Vision and Pattern Recognition (CVPR) in 2010, and a
paper that won the Prize Paper award in IEEE Transactions on Multimedia (T-MM) in 2014. He is a fellow of the IEEE.
\end{IEEEbiography}

\begin{IEEEbiography}
[{\includegraphics[width=1in,height=1.25in,clip,keepaspectratio]{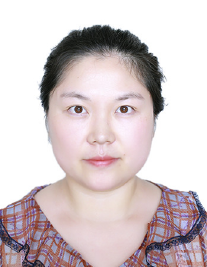}}]
{Jing Zhang} received her Ph.D. degree from University of Wollongong, Australia, in 2019. She is now a lecturer in College of Software, Beihang University, Beijing, China. Her recent research interests include machine learning and computer vision, with the special interests in transfer learning, domain adaptation, and their applications in computer vision.
\end{IEEEbiography}

\begin{IEEEbiography}
[{\includegraphics[width=1in,height=1.25in,clip,keepaspectratio]{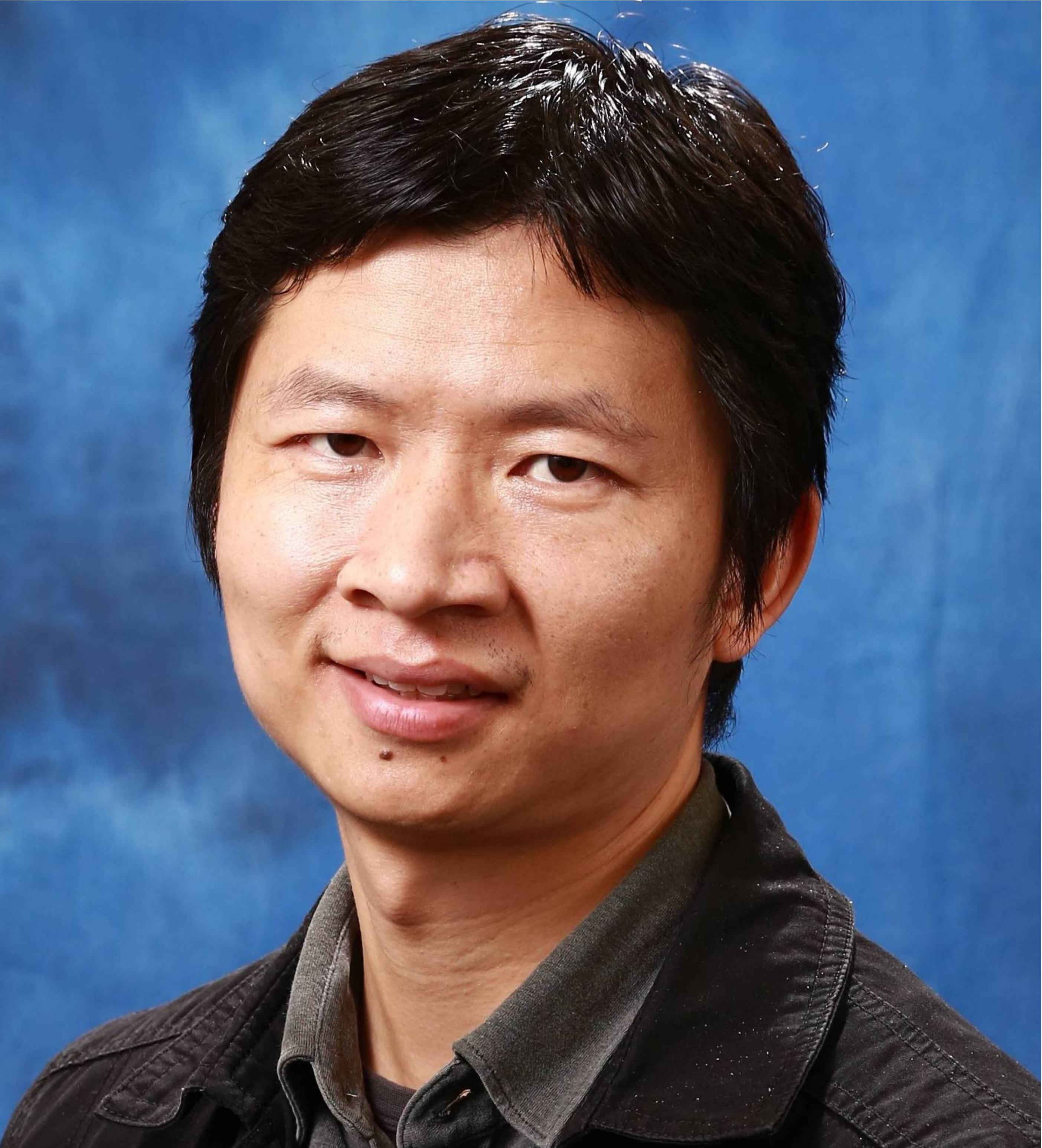}}]
{Wanli Ouyang} received the PhD degree in the Department of Electronic Engineering, Chinese University of Hong Kong. Since 2017, he has been a Senior Lecturer with The University of Sydney. His research interests include image processing, computer vision, and pattern recognition. He is a senior member of the IEEE.
\end{IEEEbiography}




\end{document}

%% file: definitions.tex
\usepackage{etoolbox}
\usepackage{bm}
\usepackage{amsthm}%
\usepackage{subfigure}%

\newcommand{\MyMapTemplatePrefixc}[4]{\expandafter#1\csname#3#4\endcsname{#2{#4}}} 
\forcsvlist{\MyMapTemplatePrefixc {\def} {\mathcal}{c}} {A,B,C,D,E,F,G,H,I,J,K,L,M,N,O,P,Q,R,S,T,U,V,W,X,Y,Z}  

\newcommand{\MyMapTemplatePrefixtb}[5]{\expandafter#1\csname#4#5\endcsname{#2{#3{#5}}}} 
\forcsvlist{\MyMapTemplatePrefixtb {\def} {\tilde}{\mathbf}{t}} {A,B,C,D,E,F,G,H,I,J,K,L,M,N,O,P,Q,R,S,T,U,V,W,X,Y,Z}  
\forcsvlist{\MyMapTemplatePrefixtb {\def} {\tilde}{\mathbf}{t}} {0,1,a,b,c,d,e,f,g,h,i,j,k,l,m,n,o,p,q,r,s,u,v,w,x,y,z}  

\newcommand{\MyMapTemplateNoPrefix}[3]{\expandafter#1\csname#3\endcsname{#2{#3}}}
\forcsvlist{\MyMapTemplateNoPrefix {\def} {\mathbf} } {0,1,a,b,c,d,e, f, g, h, i, j, k, l, m, n, o, p, q, r, u, v, w, x, y, z} 
\forcsvlist{\MyMapTemplateNoPrefix {\def} {\mathbf} } {A,B,C,D,E,F,G,H,I,J,K,L,M,N,O,P,Q,R,S,T,U,V,W,X,Y,Z}  

\def\bt{\bm{\theta}}

\def\ie{\emph{i.e.}}
\def\eg{\emph{e.g.}}
\def\resp{\emph{resp.}}

\usepackage{booktabs} 
\usepackage[table,dvipsnames]{xcolor}
\definecolor{rowblue}{RGB}{220,230,240}